\pdfoutput=1

\documentclass[11pt]{article}

\usepackage{acl}

\usepackage{times}
\usepackage{latexsym}
\usepackage{multirow}
\usepackage{subcaption}
\usepackage{bm}
\usepackage{amssymb}
\usepackage{enumitem}

\usepackage[T1]{fontenc}

\usepackage[utf8]{inputenc}
\usepackage{amsmath}

\usepackage{microtype}
\usepackage{booktabs} 
\usepackage{adjustbox}

\usepackage{inconsolata}
\usepackage{cleveref}

\newcommand\norm[1]{\left\Vert#1\right\Vert}

\title{An Unsupervised Approach to Achieve \\Supervised-Level Explainability in Healthcare Records}

\author{Joakim Edin\textsuperscript{1,3,}\thanks{ Correspondance to \url{je@corti.ai}}  \\ \bf Lasse Borgholt\textsuperscript{3} \vspace{0.2cm} \\ University of Copenhagen\textsuperscript{1} \And Maria Maistro\textsuperscript{1} \\ \bf Jakob D. Havtorn\textsuperscript{3} \vspace{0.2cm} \\ LUT University\textsuperscript{2} \And Lars Maaløe\textsuperscript{3}  \\ \bf Tuukka Ruotsalo\textsuperscript{1,2} \vspace{0.2cm} \\ Corti\textsuperscript{3}}

\begin{document}
\maketitle

\begin{abstract}
Electronic healthcare records are vital for patient safety as they document conditions, plans, and procedures in both free text and medical codes. Language models have significantly enhanced the processing of such records, streamlining workflows and reducing manual data entry, thereby saving healthcare providers significant resources.
However, the black-box nature of these models often leaves healthcare professionals hesitant to trust them. State-of-the-art explainability methods increase model transparency but rely on human-annotated evidence spans, which are costly. In this study, we propose an approach to produce plausible and faithful explanations without needing such annotations. We demonstrate on the automated medical coding task that adversarial robustness training improves explanation plausibility and introduce AttInGrad, a new explanation method superior to previous ones. By combining both contributions in a fully unsupervised setup, we produce explanations of comparable quality, or better, to that of a supervised approach. We release our code and model weights. 
\footnote{\url{https://github.com/JoakimEdin/explainable-medical-coding}}

\end{abstract}

\section{Introduction}
Explainability in natural language processing remains a largely unsolved problem, posing significant challenges for healthcare applications~\cite{lyuFaithfulModelExplanation2023}. For every patient admission, a healthcare professional must read extensive documentation in the healthcare records to assign appropriate medical codes. A code is a machine-readable identifier for a diagnosis or procedure, pivotal for tasks such as statistics, documentation, and billing. This process can involve sifting through thousands of words to choose from over 140,000 possible codes~\cite{johnsonMIMICIIIFreelyAccessible2016}, making medical coding not only time-consuming but also error-prone~\cite{burnsSystematicReviewDischarge2012, tsengAdministrativeCostsAssociated2018}.

Automated medical coding systems, powered by machine learning models, aim to alleviate these burdens by suggesting medical codes based on free-form written documentation. However, when reviewing suggested codes, healthcare professionals must still manually locate relevant evidence in the documentation. This is a slow and strenuous process, especially when dealing with extensive documentation and numerous medical codes. Explainability is essential for making this process tractable. 
\begin{figure}[t]
    \centering
    \includegraphics[width=\linewidth]{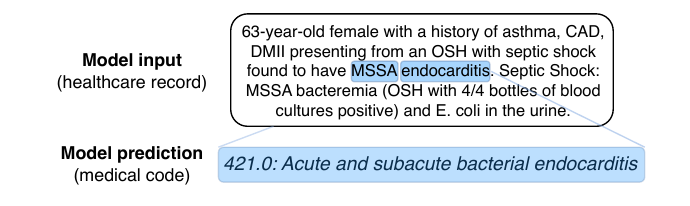}
    \caption{Example of an input, prediction, and feature attribution explanation highlighted in the input.}
    \label{fig:example}
    \vspace{-3mm}
\end{figure}

Feature attributions, a common form of explainability, can help healthcare professionals quickly find the evidence necessary to review a medical code suggestion (see \Cref{fig:example}). Feature attribution methods score each input feature based on its influence on the model's output. These explanations are often evaluated through \textit{plausibility} and \textit{faithfulness}~\cite{jacoviFaithfullyInterpretableNLP2020}. Plausibility measures how convincing an explanation is to human users, while faithfulness measures the explanation's ability to reflect the model's logic. 

While previous work has proposed feature attribution methods for automated medical coding, they only evaluated attention-based feature attribution methods. Furthermore, the state-of-the-art method uses a supervised approach relying on costly evidence-span annotations. This reliance on manual annotations significantly limits practical applicability, as each code system and its versions require separate manual annotations~\cite{mullenbachExplainablePredictionMedical2018, tengExplainablePredictionMedical2020, dongExplainableAutomatedCoding2021, kimCanCurrentExplainability2022, chengMDACEMIMICDocuments2023}.

In this study, we present an approach for producing explanations of comparable quality to the supervised state-of-the-art method but without using evidence span annotations. We implement adversarial robustness training strategies to decrease the model's dependency on irrelevant features, thereby avoiding such features in the explanations~\cite{tsiprasRobustnessMayBe2018}. Moreover, we present more faithful feature attribution methods than the attention-based method used in previous studies. Our key contributions are:
\begin{enumerate}[leftmargin=5mm]
    \itemsep0em 
    \item We show that adversarially robust models produce more plausible medical coding explanations.
    \item We propose a new feature attribution method, AttInGrad, which produces substantially more faithful and plausible medical coding explanations than previous methods.
    \item We demonstrate that the combination of an adversarial robust model and AttInGrad produces medical coding explanations of similar, or better, plausibility and faithfulness compared to the supervised state-of-the-art approach.
    \vspace{1mm}
\end{enumerate}

\section{Related work}
Next, we present explainability approaches for automated medical coding and previous work on how adversarial robustness affects explainability.

\subsection{Explainable automated medical coding}\label{seq:emc}
Automated medical coding is a multi-label classification task that aims to predict a set of medical codes from $J$ classes based on a given medical document~\cite{edinAutomatedMedicalCoding2023}. In this context, the objective of explainable automated medical coding is to generate feature attribution scores for each of the $J$ classes. These scores quantify how much each input token influences each class's prediction.

Most studies in explainable automated medical coding use attention weights as feature attribution scores without comparing to other methods~\cite{mullenbachExplainablePredictionMedical2018,tengExplainablePredictionMedical2020,dongExplainableAutomatedCoding2021, feuchtDescriptionbasedLabelAttention2021, chengMDACEMIMICDocuments2023}. However, two studies suggest alternative feature attribution methods. \citet{xuMultimodalMachineLearning2019} propose a feature attribution method tailored to their one-layer CNN architecture but do not compare performance with other methods. \citet{kimCanCurrentExplainability2022} train a linear medical coding model using knowledge distillation and use its weights as the explanation. However, their method does not improve over the explanations of the popular attention approach. \citet{chengMDACEMIMICDocuments2023} improve the plausibility of the attention weights of the final layer by training them to align with evidence span annotations. However, obtaining such annotations is costly. 

Previous work focused on plausibility using human ratings~\cite{mullenbachExplainablePredictionMedical2018, tengExplainablePredictionMedical2020, kimCanCurrentExplainability2022}, example inspection~\cite{dongExplainableAutomatedCoding2021, feuchtDescriptionbasedLabelAttention2021, liuHierarchicalLabelwiseAttention2022}, or evidence span overlap metrics~\cite{xuMultimodalMachineLearning2019, chengMDACEMIMICDocuments2023}. Notably, no studies have assessed the faithfulness of the explanations nor compared the attention-based methods with other established methods. 

\subsection{Adversarial robustness and explainability}\label{sec:adv}
Adversarial robustness refers to the ability of a machine learning model to maintain performance under adversarial attacks, which involve making small changes to input data that do not significantly affect human perception or judgment (e.g., a small amount of image noise). \citet{tsiprasRobustnessMayBe2018} and \citet{ilyasAdversarialExamplesAre2019a} demonstrate that adversarial examples exploit the models' dependence on fragile, non-robust features. Adversarial robustness training embeds invariances to prevent models relying on such non-robust features, with regularization and data augmentation as main strategies~\cite{tsiprasRobustnessMayBe2018, rosImprovingAdversarialRobustness2018}. 

Previous work in image classification shows that adversarially robust models generate more plausible explanations~\cite{rosImprovingAdversarialRobustness2018,chenRobustAttributionRegularization2019, etmannConnectionAdversarialRobustness2019a}. These studies demonstrate this phenomenon for three adversarial training strategies: 1) input gradient regularization, which improves the Lipschitzness of neural networks~\cite{druckerImprovingGeneralizationPerformance1992, rosImprovingAdversarialRobustness2018,chenRobustAttributionRegularization2019, etmannConnectionAdversarialRobustness2019a, roscaCaseNewNeural2020, felHowGoodYour2022, khanAnalyzingExplainerRobustness2023}, 2) adversarial training, which trains models on adversarial examples, thereby embeds invariance to adversarial noise~\cite{tsiprasRobustnessMayBe2018}, and 3) feature masking, which masks unimportant features during training to embed invariance to such features~\cite{bhallaDiscriminativeFeatureAttributions2023}.

The relationship between model robustness and explanation plausibility in Natural Language Processing (NLP) remains unclear, primarily due to the fundamental differences between textual tokens and image pixels. To date, only two studies, \citet{yooImprovingAdversarialTraining2021a} and \citet{liFaithfulExplanationsText2023}, have explored this relationship in depth. These studies suggest that robust models produce more faithful explanations compared to their non-robust counterparts. However, their conclusions may be questionable due to the use of the Area Over the Perturbation Curve (AOPC) metric, which is unsuitable for cross-model comparisons~\cite{edinNormalizedAOPCFixing2024a}. This inappropriate application of AOPC may have led to potentially misleading conclusions.

\section{Methods}
Here, we describe the adversarial robustness training strategies and feature attribution methods in the context of a prediction model for medical coding. The underlying automated medical coding model takes a sequence of tokens as input and outputs medical code probabilities (\Cref{sec:model}).

\subsection{Adversarial robustness training strategies}
We implemented three adversarial training strategies, which we hypothesized could decrease our medical coding model's reliance on irrelevant tokens: Input gradient regularization, projected gradient descent, and token masking. We chose these strategies because they have been shown to improve plausibility in image classification and faithfulness in text classification~\cite{liFaithfulExplanationsText2023}

\paragraph{Input gradient regularization (IGR)} encourages the gradient of the output with respect to the input to be small. This aims to decrease the number of features on which the model relies, encouraging it to ignore irrelevant words~\cite{druckerImprovingGeneralizationPerformance1992}. We adapt IGR to text classification by adding to the task's binary cross-entropy loss, $L_{\text{BCE}}$, the $\ell^2$ norm of the gradient of $L_{\text{BCE}}$ wrt. the input token embedding sequence $\bm{X}\in\mathbb{R}^{N\times D}$. This yields the total loss,
%
\begin{equation}
    L_{\text{BCE}}(f(\bm{X}),\bm{y}) + \lambda_1 \norm{ \nabla_{\bm{X}} L_{\text{BCE}}(f(\bm{X}),\bm{y})}_2 \enspace , \nonumber
\end{equation}
where $\bm{y}\in\mathbb{R}^{J}$ is a binary target vector representing the $J$ medical codes, $\lambda_1$ is a hyperparameter, and $f: \mathbb{R}^{N\times D} \rightarrow \mathbb{R}^{J}$ is the classification model.

\paragraph{Projected gradient descent (PGD)} increases model robustness by training with adversarial examples, thereby promoting invariance to such inputs~\cite{madryAdversarialRobustnessTheory2018}. We hypothesized that PGD reduces the model's reliance on irrelevant tokens, as adversarial examples often arise from the model's use of such unrobust features~\cite{tsiprasRobustnessMayBe2018}. PGD aims to find the noise $\bm{\delta}\in\mathbb{R}^{N\times D}$ that maximizes the loss $L_{\text{BCE}}(f(\bm{X}+\bm{\delta}),\bm{y})$ while satisfying the constraint $\norm{\bm{\delta}}_{\infty}\leq\epsilon$, where $\epsilon$ is a hyperparameter. PGD was originally designed for image classification; we adapted it to NLP by adding the noise to the token embeddings $\bm{X}$. We implemented PGD as follows,
\begin{equation}
    \bm{Z}^* = \underset{\bm{Z}}{\arg\max} \, L_{\text{BCE}}(f(\bm{X} + \bm{\delta(Z})),\bm{y}) \enspace , \nonumber
\end{equation}
and enforced the constraint $\norm{\bm{\delta}}_{\infty}\leq\epsilon$ by parameterizing $\bm{\delta}(\bm{Z}) = \epsilon \tanh(\bm{Z})$ and optimising $\bm{Z}\in\mathbb{R}^{N\times D}$ directly. We initialized $\bm{Z}$ with zeros. Finally, we tuned the model parameters using the following training objective:
%
%
\begin{equation}
    L_{\text{BCE}}(f(\bm{X}),\bm{y}) + \lambda_2 L_{\text{BCE}}(f(\bm{X}+\bm{\delta(Z^*)}),\bm{y}) \enspace , \nonumber 
\end{equation}
where $\lambda_2$ is a hyperparameter.

\paragraph{Token masking (TM)} teaches the model to predict accurately while using as few features as possible, thereby encouraging the model to ignore irrelevant words. TM uses a binary mask to occlude unimportant tokens and train the model to rely only on the remaining tokens~\cite{bhallaDiscriminativeFeatureAttributions2023,tomarIgnoranceBlissRobust2023}. 
Inspired by \citet{bhallaDiscriminativeFeatureAttributions2023},  we employed a two-step teacher-student approach.
We used two copies of the same model already trained on the automated medical coding task: a teacher $f_{t}$ with frozen model weights and a student $f_s$, which we fine-tuned. For each training batch, the first step was to learn a sparse mask $\bm{\hat{M}}\in[0,1]^{N \times D}$, that still provided enough information to predict the correct codes by minimizing:
\begin{equation}
    \Vert\bm{\hat{M}}\Vert_1+\beta \Vert f_s(\bm{X})-f_s(x_m(\bm{X}, \bm{\hat{M}})) \Vert_1 \enspace , \nonumber
\end{equation}
\noindent where $\beta$ is a hyperparameter and $x_m:\mathbb{R}^{N \times D} \rightarrow \mathbb{R}^{N \times D} $ is the masking function:
\begin{equation}
    x_m(\bm{X}, \bm{M}) = \bm{B} \odot (1-\bm{M}) + \bm{X} \odot \bm{M} \enspace , \nonumber
\end{equation}
where $\bm{B}\in\mathbb{R}^{N \times D}$ is the baseline input. We chose $\bm{B}$ as the token embedding representing the start token, followed by the mask token embedding repeated $N-2$ times, followed by the end token embedding. After optimization, we binarized the mask $\bm{M} = \text{round}(\bm{\hat{M}})$, where around 90\% of the features were masked.
Finally, we tuned the model $f_s$ using the following training objective:
%
%
\begin{equation}
    \resizebox{0.99\linewidth}{!}{%
        $\norm{f_s(\bm{X})-f_t(\bm{X})}_1 + \lambda_3 \norm{f_s(\bm{X})-f_s(x_m(\bm{X}, \bm{M}))}_1 \enspace , \nonumber$
    }
\end{equation}
where $\lambda_3$ is a hyperparameter.

\subsection{Feature attribution methods}
We evaluated several feature attribution methods for automated medical coding, categorizing them into three types: attention-based, gradient-based, and perturbation-based (more details in \Cref{sec:explanations}). \textbf{Attention-based} methods like Attention~\cite{mullenbachExplainablePredictionMedical2018}, Attention Rollout~\cite{abnarQuantifyingAttentionFlow2020a}, and AttGrad~\cite{serranoAttentionInterpretable2019} rely on the model's attention weights. \textbf{Gradient-based} methods such as InputXGrad~\cite{sundararajanAxiomaticAttributionDeep2017a}, Integrated Gradients (IntGrad)~\cite{sundararajanAxiomaticAttributionDeep2017a}, and Deeplift~\cite{shrikumarLearningImportantFeatures2017} use backpropagation to quantify the influence of input features on outputs. \textbf{Perturbation-based} methods, including LIME~\cite{ribeiroWhyShouldTrust2016}, KernelSHAP~\cite{lundbergUnifiedApproachInterpreting2017}, and Occlusion@1~\cite{ribeiroWhyShouldTrust2016}, measure the impact on output confidence by occluding input features.

Our investigation into feature attribution methods revealed an intriguing pattern: while individual methods often produced unreliable explanations, their shortcomings rarely overlapped. Attention-based methods and gradient-based approaches like InputXGrad frequently disagreed on which tokens were most important, yet both contributed valuable insights in different scenarios. This observation sparked a key question: could we leverage the complementary strengths of these methods to create a more robust attribution technique?

To address this question, we propose AttInGrad, a novel feature attribution method that combines Attention and InputXGrad. AttInGrad multiplies their respective attribution scores, aiming to amplify the importance of tokens deemed relevant by both methods while down-weighting those highlighted by only one or neither method.

We formalize the AttInGrad attribution scores for class $j$ using the following equation: 
\begin{equation}
    \begin{bmatrix}
        \bm{A}_{j1} \cdot \norm{\bm{X}_1 \odot \frac{\partial f_j}{\partial \bm{X}_1}(\bm{X})}_2
        \\
        \vdots
        \\
        \bm{A}_{jN} \cdot\norm{\bm{X}_N \odot\frac{\partial f_j}{\partial \bm{X}_N}(\bm{X})}_2
    \end{bmatrix} \enspace , \nonumber
\end{equation} 
\noindent where $\bm{A} \in \mathbb{R}^{J \times N}$ is the attention matrix, $\odot$ is the element-wise matrix multiplication operation, $N$ are the number of tokens in a document, and $J$ is the number of classes.

In \Cref{sec:special_token_analysis}, we will provide an in-depth analysis of the mechanisms underlying AttInGrad's effectiveness, shedding light on why this combination of methods yields improved feature attributions.
%

   

\section{Experimental setup}
In the following, we present our datasets, models, and evaluation metrics.

\subsection{Data}

We conducted our experiments using the open-access MIMIC-III and the newly released MDACE dataset \cite{johnsonMIMICIIIFreelyAccessible2016, chengMDACEMIMICDocuments2023}. MIMIC-III\footnote{We decided to use MIMIC-III instead of the newer MIMIC-IV because we wanted to use the same dataset as \citet{chengMDACEMIMICDocuments2023}.} includes 52,722 discharge summaries from the Beth Israel Deaconess Medical Center's ICU, collected between 2008 and 2016 and annotated with ICD-9 codes. MDACE comprises 302 reannotated MIMIC-III cases, adding evidence spans to indicate the textual justification for each medical code. Not all possible evidence spans are annotated; for example, if hypertension is mentioned multiple times, only the first mention might be annotated, leaving subsequent mentions unannotated. We focused exclusively on discharge summaries, as most previous medical coding studies on MIMIC-III \cite{tengReviewDeepNeural2022}. Statistics are in \Cref{tab:data_stats}.

For dataset splits, we used MIMIC-III full, a popular split by \citet{mullenbachExplainablePredictionMedical2018}, and MDACE, introduced by \citet{chengMDACEMIMICDocuments2023} for training and evaluating explanation methods. All MDACE examples are from the MIMIC-III full test set, which we excluded from this test set when using MDACE in our training data.

\begin{table}
    \centering
    \caption{The two data splits used in this paper. MIMIC-III full comprises discharge summaries annotated with ICD-9 codes. MDACE comprises discharge summaries annotated with ICD-9 codes and evidence spans.}
    \label{tab:splits}
    \begin{tabular}{cccc}
    \toprule
         Split & Train & Val & Test \\
         \midrule
         MIMIC-III full & 47,719 & 1,631 & 3,372 \\
         MDACE & 181 & 60 & 61 \\
         \bottomrule
    \end{tabular}
    \label{tab:data_stats}
    \vspace{-2mm}
\end{table}

\subsection{Models}\label{sec:model}

We used PLM-ICD, a state-of-the-art automated medical coding model architecture, for our experiments because its architecture is simple while outperforming other models according to ~\citet{huangPLMICDAutomaticICD2022, edinAutomatedMedicalCoding2023}. To address stability issues caused by numerical overflow in the decoder of the original model, we replaced the label-wise attention mechanism with standard cross-attention~\cite{vaswaniAttentionAllYou2017}. This adjustment not only stabilized training but also slightly improved performance. We provide further details on the architecture modifications in \Cref{app:model}.

We compared five models: $\text{B}_{\text{U}}$, $\text{B}_{\text{S}}$, IGR, PGD, and TM. All models used our modified PLM-ICD architecture but were trained differently. $\text{B}_{\text{U}}$ was trained unsupervised with binary cross-entropy, whereas $\text{B}_{\text{S}}$ employed a supervised auxiliary training objective that minimized the KL divergence between the model’s cross-attention weights and annotated evidence spans, as per \citet{chengMDACEMIMICDocuments2023}. IGR, PGD, and TM training is as in \Cref{sec:adv}. Best hyperparameters are in \Cref{sec:hyperparameter}.
\begin{figure*}[t]
    \centering
    \begin{subfigure}{0.325\textwidth}
        \centering
        \includegraphics[width=\textwidth]{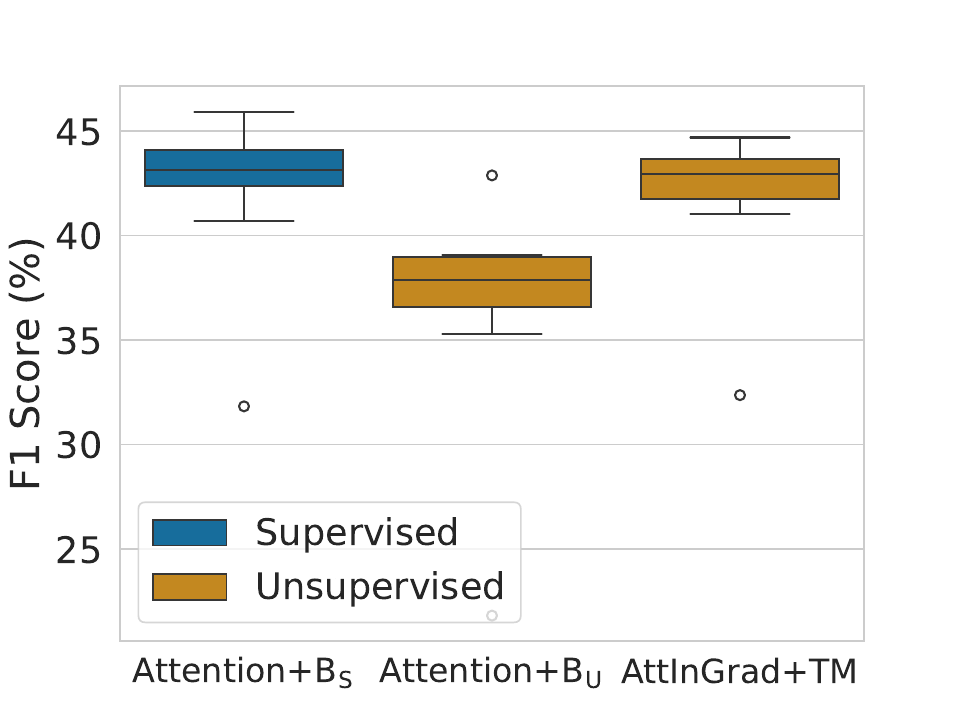}
        \caption{F1 score $\uparrow$}
    \end{subfigure}
    \begin{subfigure}{0.325\textwidth}
        \centering
        \includegraphics[width=\textwidth]{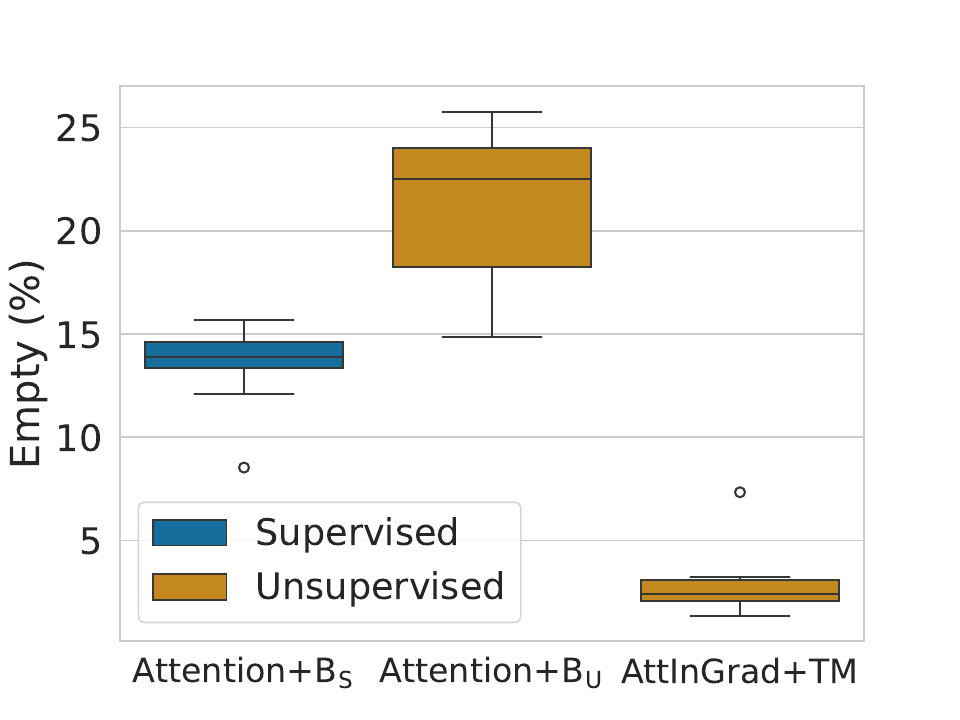}
        \caption{Empty $\downarrow$}
    \end{subfigure}
    \begin{subfigure}{0.325\textwidth}
        \centering
        \includegraphics[width=\textwidth]{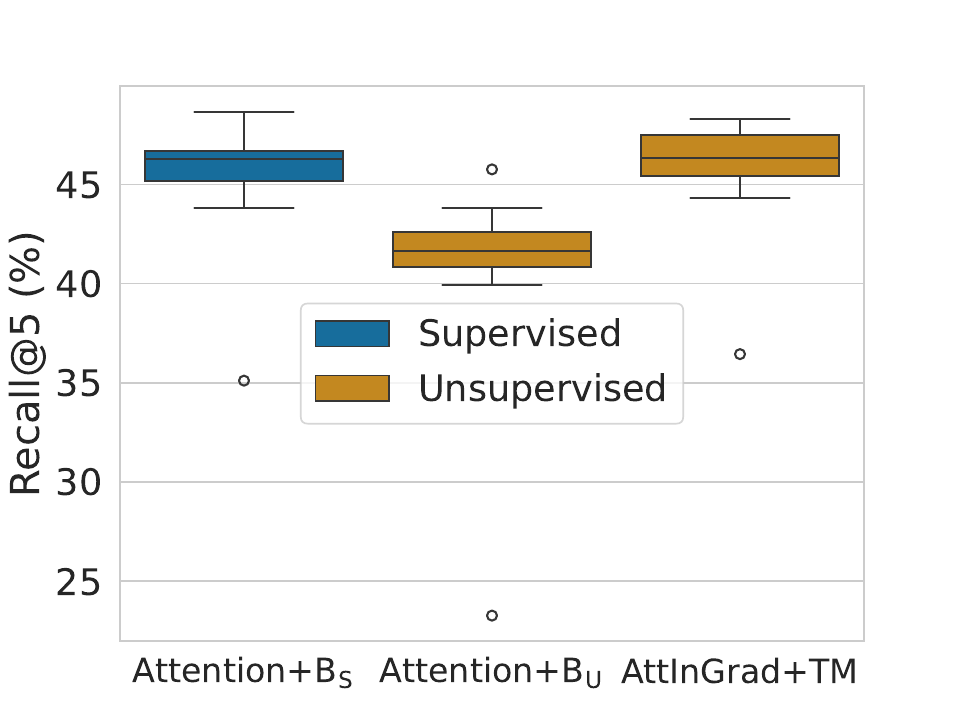}
        \caption{Recall@5 $\uparrow$}
    \end{subfigure}
    \caption{Comparison of plausibility across various combinations of explanation methods and models from this study and previous work. Most previous studies used Attention and a standard medical coding model (B$_{\text{U}}$).  \citet{chengMDACEMIMICDocuments2023} instead used a supervised model trained on evidence-span annotations (B$_{\text{S}}$). We proposed AttInGrad and an adversarial robust model (TM).}
    \label{fig:plausibility_comp}
    \vspace{-2mm}
\end{figure*}
\subsection{Experiments}

We trained all five models with ten seeds on the MIMIC-III full and MDACE training set. The supervised training strategy $\text{B}_{\text{S}}$ used the evidence span annotations, while the others only used the medical code annotations. For each model, we evaluated the plausibility and faithfulness of the explanations generated by every explanation method. 

We aimed to demonstrate a similar explanation quality as a supervised approach but without training on evidence spans. Therefore, after evaluating the models and explanation methods, we compared our best combination with the supervised strategy proposed by \citet{chengMDACEMIMICDocuments2023}, who used the B$_{\text{S}}$ model and the Attention explanation method. We also compared our best combination with the unsupervised strategy used by most previous works (see \Cref{seq:emc}), comprising the B$_{\text{U}}$ model and the Attention explanations method.

\begin{table*}[t]
\centering

\caption{Plausibility of Attention, InputXGrad, Integrated Gradients (IntGrad), Deeplift, and AttInGrad on the MDACE test set. Each experiment was run with ten different seeds. We show the mean of the seeds $\pm$ as the standard deviation. All the scores are presented as percentages. Bold numbers outperform the unsupervised baseline model, while underlined numbers outperform the supervised model. We included more feature attribution methods in \Cref{sec:full_results}.}
\label{tab:plausibility_both_sup}
\resizebox{\textwidth}{!}{%
\begin{tabular}{llcccccccccccc}
\toprule
 &  & \multicolumn{7}{c}{Prediction} & \multicolumn{3}{c}{Ranking}  \\
 \cmidrule(lr){3-9} \cmidrule(lr){10-12}
 Explainer& Model & P $\uparrow$ & R $\uparrow$ & F1 $\uparrow$ & AUPRC $\uparrow$ & Empty $\downarrow$ & SpanR $\uparrow$ & Cover $\uparrow$  & IOU $\uparrow$ & P@5 $\uparrow$ & R@5 $\uparrow$  \\ 
\midrule
\multirow[c]{4}{*}{Attention} & B$_{\text{S}}$ & 41.7$\pm$5.0 & 42.9$\pm$3.2 & 42.2$\pm$3.7 & 37.3$\pm$3.9 & 13.5$\pm$1.9 & 60.5$\pm$3.7 & 72.3$\pm$3.5 & 46.1$\pm$4.5 & 32.3$\pm$2.6 & 45.3$\pm$3.7 \\
 & B$_{\text{U}}$ & 35.9$\pm$5.9 & 37.7$\pm$5.7 & 36.6$\pm$5.3 & 31.4$\pm$6.4 & 21.1$\pm$3.6 & 53.2$\pm$7.4 & 63.4$\pm$6.8 & 39.3$\pm$6.9 & 28.8$\pm$4.2 & 40.3$\pm$5.9 \\
 & IGR & 35.5$\pm$6.2 & \textbf{40.4$\pm$6.1} & \textbf{37.7$\pm$5.9} & \textbf{32.4$\pm$6.8} & \textbf{19.4$\pm$2.1} & \textbf{55.5$\pm$8.0} & \textbf{64.5$\pm$6.2} & \textbf{40.0$\pm$7.4} & \textbf{29.2$\pm$4.5} & \textbf{40.9$\pm$6.3} \\
 & TM & \textbf{36.5$\pm$5.9} & \textbf{37.8$\pm$6.1} & \textbf{37.0$\pm$5.5} & \textbf{31.7$\pm$6.7} & 21.7$\pm$3.3 & \textbf{53.3$\pm$7.7} & 63.3$\pm$6.9 & \textbf{40.1$\pm$7.3} & \textbf{29.1$\pm$4.4} & \textbf{40.8$\pm$6.2} \\
  & PGD & 33.0$\pm$8.4 & \textbf{38.4$\pm$8.0} & 35.4$\pm$8.3 & 30.0$\pm$9.1 & \textbf{19.1$\pm$1.6} & 51.5$\pm$13.6 & 61.1$\pm$9.9 & 35.9$\pm$10.8 & 27.7$\pm$6.3 & 38.9$\pm$8.8 \\
 \midrule
\multirow[c]{4}{*}{InputXGrad} & B$_{\text{S}}$ & 30.7$\pm$2.0 & 33.7$\pm$2.6 & 32.0$\pm$1.4 & 24.9$\pm$2.0 & 7.3$\pm$1.3 & 48.5$\pm$2.8 & 64.0$\pm$2.8 & 32.2$\pm$2.1 & 26.6$\pm$1.2 & 37.3$\pm$1.6 \\
 & B$_{\text{U}}$ & 31.0$\pm$2.6 & 32.7$\pm$3.2 & 31.6$\pm$1.2 & 24.8$\pm$1.9 & 9.5$\pm$2.6 & 46.4$\pm$3.6 & 62.3$\pm$3.5 & 31.7$\pm$1.8 & 26.3$\pm$0.9 & 36.9$\pm$1.3 \\
 & IGR & \underline{\textbf{32.4$\pm$2.5}} & \underline{\textbf{33.8$\pm$2.2}} & \underline{\textbf{33.0$\pm$0.7}} & \underline{\textbf{27.0$\pm$0.7}} & 9.7$\pm$2.0 & \textbf{48.5$\pm$2.6} & \underline{\textbf{64.6$\pm$2.5}} & \underline{\textbf{32.8$\pm$1.0}} & \underline{\textbf{27.4$\pm$0.4}} & \underline{\textbf{38.4$\pm$0.5}} \\
 & TM & \underline{\textbf{32.4$\pm$2.7}} & \underline{\textbf{34.1$\pm$2.1}} & \underline{\textbf{33.1$\pm$1.4}} & \underline{\textbf{26.2$\pm$2.3}} & \textbf{8.6$\pm$1.6} & \textbf{48.5$\pm$2.4} & \underline{\textbf{64.8$\pm$2.5}} & \underline{\textbf{32.6$\pm$1.5}} & \underline{\textbf{27.4$\pm$1.1}} &\underline{\textbf{ 38.3$\pm$1.5}} \\
  & PGD & 30.1$\pm$2.2 & \textbf{32.9$\pm$1.7} & 31.4$\pm$1.4 & 24.8$\pm$1.5 & \textbf{9.3$\pm$2.2} & \textbf{46.6$\pm$2.1} & \textbf{62.6$\pm$2.6} & 31.4$\pm$1.3 & 26.1$\pm$1.0 & 36.6$\pm$1.4 \\
 \midrule
\multirow[c]{4}{*}{IntGrad} & B$_{\text{S}}$ & 30.9$\pm$3.2 & 33.9$\pm$2.0 & 32.2$\pm$1.7 & 26.2$\pm$2.6 & 4.8$\pm$2.0 & 50.8$\pm$2.4 & 65.9$\pm$3.3 & 34.2$\pm$2.8 & 27.4$\pm$1.6 & 38.4$\pm$2.2 \\
 & B$_{\text{U}}$ & 31.3$\pm$4.3 & 32.8$\pm$2.7 & 31.9$\pm$3.2 & 26.0$\pm$4.3 & 5.2$\pm$1.0 & 49.4$\pm$3.0 & 64.4$\pm$3.1 & 33.9$\pm$4.2 & 26.5$\pm$2.5 & 37.1$\pm$3.6 \\
 & IGR & \underline{\textbf{32.6$\pm$2.2}} & \textbf{33.8$\pm$2.3} & \underline{\textbf{33.1$\pm$1.3}} & \underline{\textbf{26.6$\pm$2.0}} & 5.2$\pm$1.7 & \textbf{49.7$\pm$2.5} & \textbf{64.7$\pm$3.3} & \underline{\textbf{34.9$\pm$2.3}} & \underline{\textbf{27.6$\pm$0.9}} & \underline{\textbf{38.7$\pm$1.2}} \\
 & TM & \underline{\textbf{33.1$\pm$3.6}} & \underline{\textbf{34.0$\pm$3.7}}& \underline{\textbf{33.4$\pm$2.8}} & \underline{\textbf{27.5$\pm$3.7}} & 5.4$\pm$1.6 & \underline{\textbf{51.0$\pm$3.9}} & \textbf{65.5$\pm$4.1} & \underline{\textbf{35.3$\pm$4.2}} & \underline{\textbf{27.6$\pm$2.1}} & \underline{\textbf{38.7$\pm$3.0}} \\
  & PGD & 30.0$\pm$5.4 & \textbf{33.5$\pm$3.3} & 31.4$\pm$4.2 & 25.4$\pm$4.8 & \textbf{5.1$\pm$1.8} & \textbf{49.7$\pm$3.6} & \textbf{64.5$\pm$3.9} & 33.6$\pm$4.8 & 26.4$\pm$3.0 & 36.9$\pm$4.2 \\
 \midrule
\multirow[c]{4}{*}{Deeplift} & B$_{\text{S}}$ & 29.2$\pm$2.3 & 34.2$\pm$3.0 & 31.3$\pm$1.2 & 24.1$\pm$1.8 & 6.4$\pm$1.8 & 48.9$\pm$3.4 & 65.2$\pm$3.3 & 31.1$\pm$1.9 & 26.2$\pm$1.1 & 36.7$\pm$1.5 \\
 & B$_{\text{U}}$ & 31.2$\pm$1.9 & 31.4$\pm$2.4 & 31.2$\pm$1.5 & 24.1$\pm$2.1 & 9.1$\pm$1.6 & 45.1$\pm$2.5 & 61.7$\pm$2.9 & 30.9$\pm$1.6 & 25.8$\pm$1.1 & 36.1$\pm$1.5 \\
 & IGR & \underline{\textbf{31.0$\pm$1.8}} & \textbf{33.7$\pm$1.4} & \underline{\textbf{32.2$\pm$0.6}} & \underline{\textbf{25.9$\pm$0.9}} & \textbf{8.6$\pm$1.0} & \textbf{48.3$\pm$1.5} & \textbf{64.8$\pm$1.7} & \underline{\textbf{31.6$\pm$1.0}} & \underline{\textbf{26.7$\pm$0.5}} & \underline{\textbf{37.4$\pm$0.7}} \\
 & TM & \underline{\textbf{30.2$\pm$2.4}} & \underline{\textbf{34.8$\pm$1.4}} & \underline{\textbf{32.2$\pm$1.4}} & \underline{\textbf{25.1$\pm$2.5}} & \textbf{6.8$\pm$1.5} & \underline{\textbf{49.1$\pm$1.8}} & \underline{\textbf{65.7$\pm$1.6}} & \underline{\textbf{31.4$\pm$1.8}} & \underline{\textbf{26.6$\pm$1.1}} & \underline{\textbf{37.3$\pm$1.6}} \\
  & PGD & 29.4$\pm$2.8 & \textbf{32.8$\pm$1.5} & 30.9$\pm$1.5 & 23.9$\pm$1.5 & \textbf{8.1$\pm$1.8} & \textbf{46.7$\pm$1.}8 & \textbf{63.3$\pm$2.2} & 30.7$\pm$1.4 & 25.5$\pm$0.9 & 35.8$\pm$1.2 \\
 \midrule

\multirow[c]{4}{*}{AttInGrad} & B$_{\text{S}}$ & 40.6$\pm$2.2 & 45.9$\pm$3.1 & 43.0$\pm$1.9 & 38.8$\pm$2.1 & 1.2$\pm$0.5 & 63.9$\pm$2.9 & 79.0$\pm$1.9 & 45.7$\pm$2.6 & 33.3$\pm$1.4 & 46.6$\pm$1.9 \\
 & B$_{\text{U}}$ & 40.7$\pm$2.9 & 42.5$\pm$4.4 & 41.5$\pm$3.2 & 37.1$\pm$3.6 & 3.0$\pm$1.3 & 59.3$\pm$5.4 & 74.9$\pm$5.2 & 43.5$\pm$4.0 & 32.0$\pm$2.1 & 44.9$\pm$3.0 \\
 & IGR & 38.8$\pm$4.5 & \textbf{44.0$\pm$4.0} & 41.2$\pm$4.0 & \textbf{37.2$\pm$4.5} & \textbf{2.3$\pm$0.7} & \textbf{60.3$\pm$5.3} & 74.5$\pm$4.6 & 43.3$\pm$4.9 & 31.9$\pm$2.7 & 44.7$\pm$3.8 \\
 & TM & 40.2$\pm$3.0 & \textbf{43.9$\pm$4.8} & \textbf{41.9$\pm$3.4} & \textbf{37.8$\pm$3.9} & \textbf{2.8$\pm$1.6} & \textbf{60.5$\pm$5.8} & \textbf{75.9$\pm$5.3} & \textbf{44.0$\pm$3.9} & \textbf{32.5$\pm$2.3} & \textbf{45.5$\pm$3.2} \\
 & PGD & 37.1$\pm$8.2 & 42.5$\pm$5.7 & 39.3$\pm$7.1 & 34.7$\pm$8.6 & \textbf{2.3$\pm$0.9} & 57.7$\pm$9.1 & 72.7$\pm$7.4 & 39.6$\pm$10.2 & 30.9$\pm$4.8 & 43.3$\pm$6.7 \\
\bottomrule
\end{tabular}}
\end{table*}

\subsection{Evaluation metrics}\label{sec:eval}
We measured the explanation quality using metrics estimating plausibility and faithfulness. Plausibility measures how convincing an explanation is to human users, while faithfulness measures how accurate an explanation reflects a model's true reasoning process~\cite{jacoviFaithfullyInterpretableNLP2020}.

\paragraph{Plausibility metrics}
Our plausibility metrics measured the overlap between explanations and annotated evidence-spans. We assumed that a high overlap indicated plausible explanations for medical coders. We identified the most important tokens using feature attribution scores, applying a decision boundary for classification metrics, and selecting the top $K$ scores for ranking metrics.

For classification metrics, we used Precision (P), Recall (R), and F1 scores, selecting the decision boundary that yielded the highest F1 score on the validation set~\cite{chengMDACEMIMICDocuments2023}. Additionally, we included four more classification metrics: Empty explanation rate (Empty), Evidence span recall (SpanR), Evidence span cover (Cover), and Area Under the Precision-Recall Curve (AUPRC). Empty measures the rate of empty explanations when all attribution scores in an example are below the decision boundary. SpanR measures the percentage of annotated evidence spans where at least one token is classified correctly. Cover measures the percentage of tokens in an annotated evidence span that are classified correctly, given that at least one token is predicted correctly. AUPRC represents the area under the precision-recall curve generated by varying the decision boundary from zero to one.

For ranking metrics, we selected the top $K$ tokens with the highest attribution scores, using Recall@K, Precision@K, and Intersection-Over-Unions (IOU) \cite{deyoungERASERBenchmarkEvaluate2020}.

\paragraph{Faithfulness metrics}

We use two metrics to approximate faithfulness: Sufficiency and Comprehensiveness~\cite{deyoungERASERBenchmarkEvaluate2020}; more details are in \Cref{app:faithful}. Faithful explanations yield high Comprehensiveness and low Sufficiency scores. A high Sufficiency score indicates that many important tokens are incorrectly assigned low attribution scores, while a low Comprehensiveness score suggests that many non-important tokens are incorrectly assigned high attribution scores.

\section{Results}
Next, we present experimental results for the different training strategies and explainability methods.

\begin{figure}[t]
    \centering
    \begin{subfigure}{0.9\linewidth}
        \includegraphics[width=\textwidth]{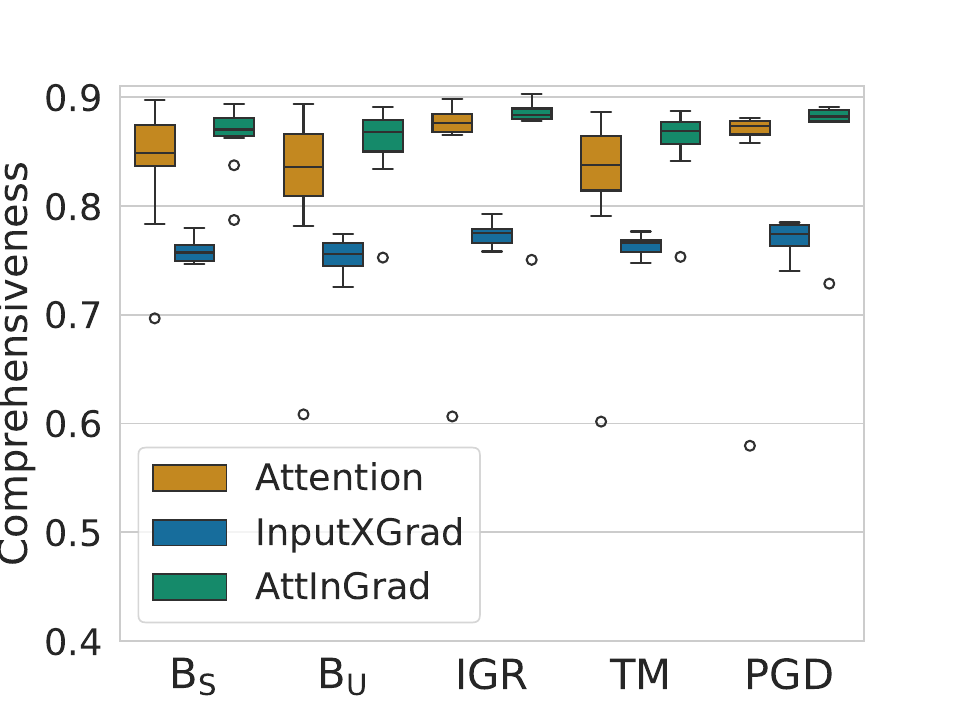}
        \caption{Comprehensiveness $\uparrow$}
    \end{subfigure}
    \begin{subfigure}{0.9\linewidth}
        \includegraphics[width=\textwidth]{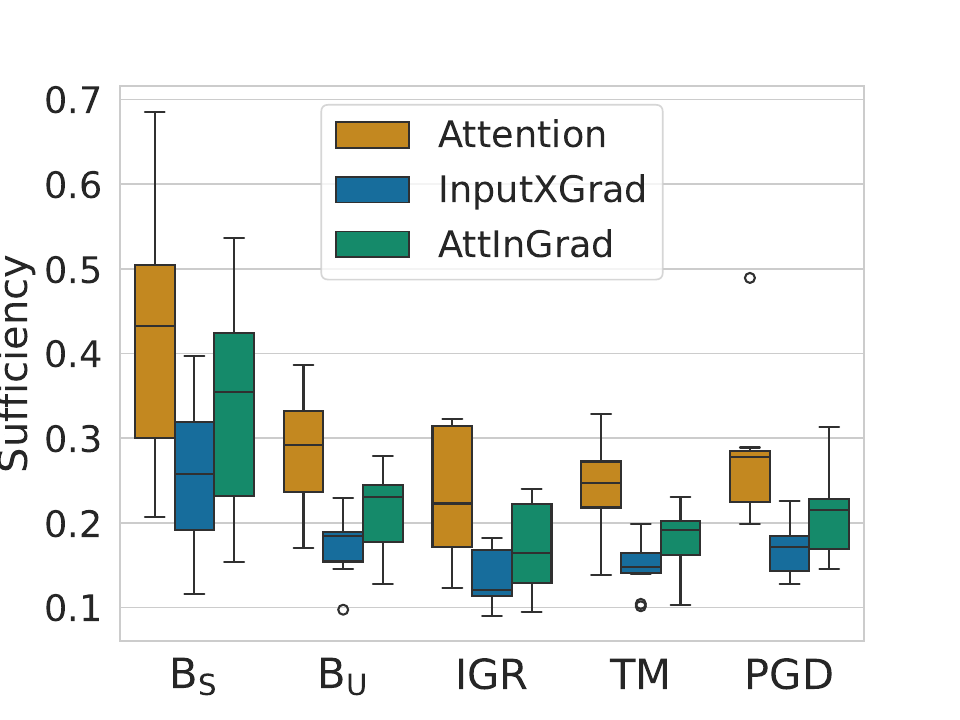}
        \caption{Sufficiency $\downarrow$}
    \end{subfigure}
    \caption{Faithfulness of Attention, InputXGrad, and AttInGrad across models.}
    \label{fig:faitfulness_boxplot_sup}
    \vspace{-2mm}
\end{figure}

\paragraph{Rivaling supervised methods in explanation quality}\label{sec:sup_res}
The objective of this paper was to produce high-quality explanations without relying on evidence span annotations. In \Cref{fig:plausibility_comp}, we compare the plausibility of our approach (Token masking and AttnInGrad) with the unsupervised approach (B$_{\text{U}}$ and Attention) and supervised state-of-the-art approach (B$_{\text{S}}$ and Attention). 
Our approach was substantially more plausible than the unsupervised on all metrics. Compared with the supervised, our approach achieved similar F1 and Recall@5 and substantially better Empty scores. 
The supervised approach achieved similar plausibility to ours on most metrics (see \Cref{tab:plausibility_both_sup}).
Our approach also achieved the highest comprehensiveness and lowest sufficiency scores (see \Cref{fig:faitfulness_boxplot_sup}). The difference was larger in the sufficiency scores, where the supervised score was twice as high as ours.

\paragraph{Adversarial robustness improves plausibility}

We evaluated the explanation plausibility of every model and explanation method combination in \Cref{tab:plausibility_both_sup}. IGR and TM outperformed the baseline model $\text{B}_{\text{U}}$ on most metrics and explanation methods. In \Cref{app:unsupervised}, we compare the unsupervised models on a bigger test set and see similar results. 
The supervised model $\text{B}_{\text{S}}$ yielded better results for attention-based explanations but was weaker than the robust models when using the gradient-based explanation methods: InputXGrad, IG, and Deeplift. 

\paragraph{AttInGrad is more plausible and faithful than Attention}\label{sec:att_plaus_faith}

AttInGrad was more plausible than all other explanation methods across all training strategies and metrics. Notably, plausibility improvements were particularly significant, with relative gains exceeding ten percent in most metrics (see \Cref{tab:plausibility_both_sup} and \Cref{sec:full_results}). For instance, for B$_{\text{U}}$, AttInGrad reduced the Empty metric from 21.1\% to 3.0\% and improved the Cover metric from 63.4\% to 74.9\%. However, these enhancements were less pronounced for B$_{\text{S}}$, the supervised model.

AttInGrad was also more faithful than Attention (see \Cref{fig:faitfulness_boxplot_sup}). However, while AttInGrad surpassed the gradient-based methods in comprehensiveness, its sufficiency scores were slightly worse.

\paragraph{Analysis of attention-based explanations}\label{sec:special_token_analysis}
While AttInGrad and Attention were more plausible than the gradient-based explanations, they had a three-fold higher inter-seed variance. We found that they often attributed high importance to tokens devoid of alphanumeric characters such as Ġ[, *, and Ċ, which we classify as \textit{special tokens}. These special tokens, such as punctuation and byte-pair encoding artifacts, rarely carry semantic meaning. In the MDACE test set, they accounted for 32.2\% of all tokens, compared to just 5.8\% within the annotated evidence spans, suggesting they are unlikely to be relevant evidence.

\begin{figure}[t]
    \centering
    \begin{subfigure}{0.9\linewidth}
    \centering
        \includegraphics[width=\textwidth]{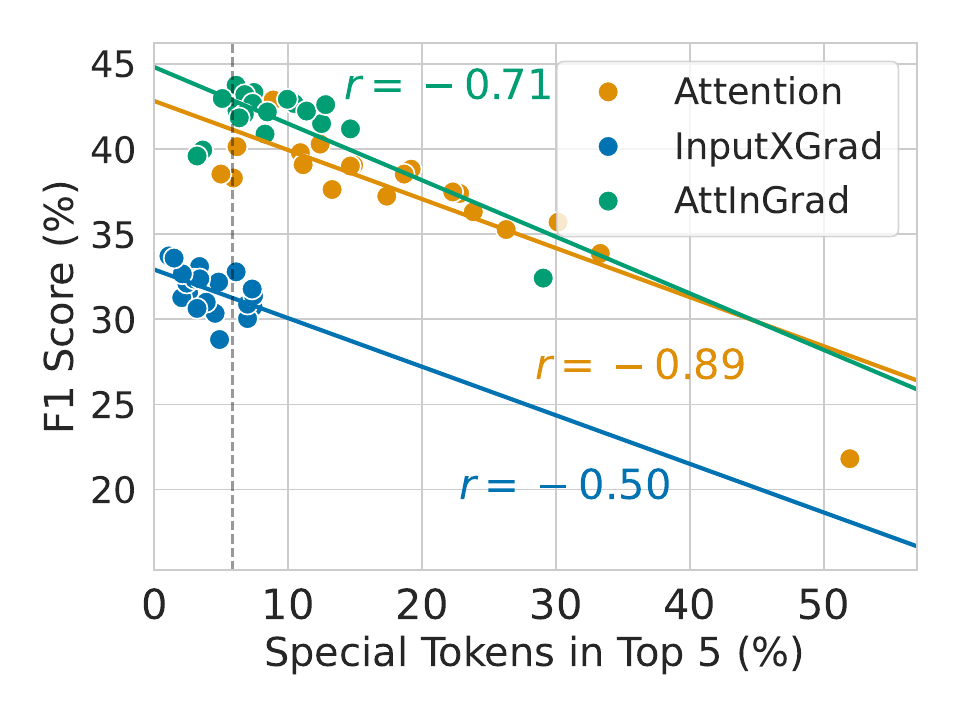}
        \caption{Plausibility F1 score $\uparrow$}
        \label{fig:special_tokens_f1}
    \end{subfigure}
    \begin{subfigure}{0.9\linewidth}
    \centering
        \includegraphics[width=\textwidth]{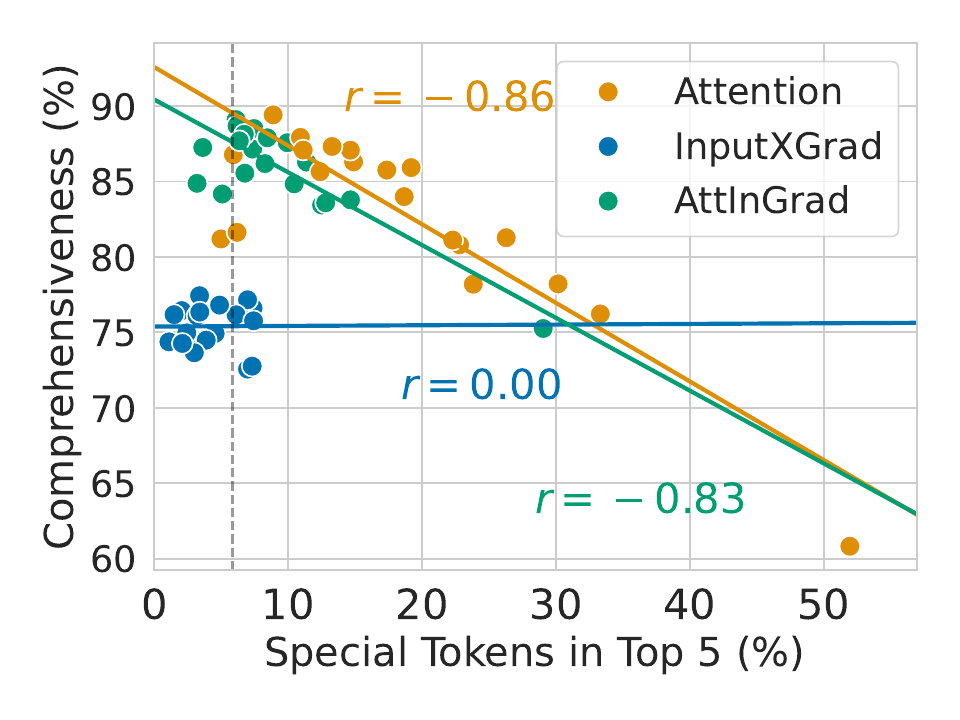}
        \caption{Comprehensiveness $\uparrow$}
        \label{fig:special_tokens_comp}
    \end{subfigure}
    \caption{The relationship between explanation quality and the proportion of the top five most important tokens that are special tokens (tokens devoid of alphanumeric characters). Each data point is the average statistic on the MDACE test set for a seed of B$_{\text{U}}$. We fitted a linear regression for each explanation method and calculated the Pearson correlation ($r$). The dotted vertical lines represent the proportion of special tokens in the evidence-span annotations.}
    \label{fig:special_tokens}
    \vspace{-2mm}
\end{figure}

In \Cref{fig:special_tokens}, we analyze the relationship between explanation quality (y-axis) and the proportion of the top five most important tokens that are special tokens (x-axis). Each data point represents the average statistics across the MDACE test set for one seed/run of the B$_\text{U}$ model. Figures~\ref{fig:special_tokens_f1} and~\ref{fig:special_tokens_comp} show F1 (plausibility) and comprehensiveness (faithfulness) respectively.
For Attention and AttInGrad, we see strong negative correlations for both metrics with a large inter-seed variance. The regression lines fitted on Attention and AttInGrad overlap, with the data points from AttInGrad shifted slightly towards the upper left, indicating attribution of less importance to special tokens.

Conversely, for InputXGrad, we see a moderate negative correlation for the F1 score and no correlation for comprehensiveness. Furthermore, InputXGrad demonstrates a small inter-seed variance, where the proportion of special tokens more closely mirrors that observed in the evidence spans. 

We hypothesized that AttInGrad's improvements over Attention stem from InputXGrad reducing special tokens' attribution scores. We tested this by zeroing out these tokens' scores. While it substantially enhanced Attention's F1 score, Attention remained lower than AttInGrad (see \Cref{tab:special_tokens}). If AttInGrad's sole contribution were filtering special tokens, we would expect similar F1 scores after zeroing their attributions. The fact that AttInGrad still outperforms Attention after controlling for special tokens suggests that there are additional factors beyond special token filtering contributing to AttInGrad's improved performance. 

\section{Discussion}

\paragraph{Do we need evidence span annotations?}
We demonstrated that we could match the explanation quality of \citet{chengMDACEMIMICDocuments2023} but without supervised training with evidence-span annotations (see \Cref{sec:sup_res}). This raises the question: are evidence-span annotations unnecessary?

Intuitively, training a model on evidence spans should encourage it to use features relevant to humans, thereby making its explanations more plausible. However, we hypothesize that the training strategy used by \citet{chengMDACEMIMICDocuments2023} primarily addresses the shortcomings of attention-based explanation methods rather than enhancing the model's underlying logic. The model B$_{\text{S}}$ only produced more plausible explanations with attention-based feature attribution methods (see \Cref{fig:plausibility_comp}). If the model truly leveraged more informative features, we would expect to see improvements across various feature attribution methods. Additionally, the differences between Attention and AttInGrad were negligible for  B$_{\text{S}}$ compared to the other models.
This may suggest that the supervised training might have corrected some of the inherent issues in the Attention method, similar to what AttInGrad achieves.

\paragraph{Adversarial robustness training strategies' impact on explanation plausibility}

While IGR and TM generated more plausible explanations than B$_{\text{U}}$, our evidence is insufficient to conclude whether the improvements were caused by our adversarially robust models relying on fewer irrelevant features. The adversarial robustness training strategies, especially PGD, had a larger impact on the plausibility of the explanations in previous image classification studies~\cite{tsiprasRobustnessMayBe2018}. We speculate that this discrepancy is caused by the inherent differences in the text and image modalities, causing techniques designed for image classifiers to be less effective for text classifiers~\cite{etmannConnectionAdversarialRobustness2019a}. 


\begin{table}[t]
    \centering
    \caption{Impact on F1 score of zeroing out feature attribution scores of special tokens for B$_{\text{U}}$.}
    \label{tab:special_tokens}

    \begin{tabular}{ccc}
    \toprule
     & Before & After \\
     \midrule
        Attention  & 36.5$\pm$5.4  & 40.2$\pm$3.9\\
        InputXGrad & 31.6$\pm$1.2 & 31.7$\pm$1.2\\
        AttInGrad & 41.5$\pm$4.4 & 42.4$\pm$2.4\\
    \bottomrule
    \end{tabular}

    \vspace{-2mm}
\end{table}

\paragraph{Limitations of attention-based explanations}
Despite Attention and AttInGrad outperforming other methods in plausibility and faithfulness, they exhibited significant shortcomings, including high sufficiency and inter-seed variation. These findings align with previous research questioning the faithfulness of solely relying on final layer attention weights~\cite{jainAttentionNotExplanation2019}.

We hypothesize these limitations stem from misalignment between the positions of the original tokens and their encoded representations. Our analysis (\Cref{sec:special_token_analysis}) suggests the encoder may store contextual information in uninformative tokens, such as special tokens, which are then used by the final attention layer for classification. As the training loss does not penalize where contextualized information is placed, this location can vary across training iterations, leading to the observed high inter-seed variance in attention-based explanations.

Training strategies that enforce alignment between original tokens and their encoded representations could alleviate the limitations of Attention and AttInGrad. This alignment might explain the benefits of the supervised training strategy proposed by \citet{chengMDACEMIMICDocuments2023}. However, rather than restricting the model, future research should explore feature attribution methods that incorporate information from all transformer layers, not just the final one~\cite{kobayashiIncorporatingResidualNormalization2021}. Although attention rollout, a method incorporating all attention layers, proved unsuccessful in our experiments (see \Cref{sec:full_results}), recent studies have highlighted its shortcomings and proposed alternative feature attribution methods that may be more suitable for our task~\cite{modarressiGlobEncQuantifyingGlobal2022a, modarressiDecompXExplainingTransformers2023a}.

\paragraph{Recommendations}
Similar to \citet{lyuFaithfulModelExplanation2023}, we advocate that future research on feature attribution methods prioritize enhancing their faithfulness, as focusing solely on plausibility can yield misleading explanations. When models misclassify or rely on irrelevant features, explanations can only appear plausible if they ignore the model's actual reasoning process. Overemphasizing plausibility may inadvertently lead researchers to favor approaches that produce explanations disconnected from the model’s true reasoning.

Instead, we propose that researchers prioritize improving the faithfulness of feature attribution methods while also working to align the model's reasoning process with that of humans. This approach not only enhances the plausibility and faithfulness of explanations but also contributes to the accuracy and robustness of model classifications.

\section{Conclusion}
Our goal was to enhance the plausibility and
the faithfulness of explanations without evidence-span annotations. We found that training our model using input gradient regularization or token masking resulted in more plausible gradient-based explanations. We proposed a new explanation method, AttInGrad, which was substantially more plausible and faithful than the attention-based explanation method used in previous studies. By combining the best training strategy and explanation method, we showed results of similar quality to a supervised baseline \cite{chengMDACEMIMICDocuments2023}. 

\section*{Limitations}
Our study did not conclusively show why adversarial robustness training strategies improved the explanation plausibility. We hypothesized that these strategies force the model to rely on fewer features that weakly correlate with the labels, and such features are less plausible. However, validating this hypothesis proved challenging. Our analysis of feature attributions' entropy was inconclusive, as detailed in \Cref{sec:entropy}. Moreover, we did not know which features the model relied on because this would require a perfect feature attribution method, which is what we aimed to develop. Despite these challenges, we demonstrated that the adversarial robust models produced more plausible explanations. We believe that our work has laid a solid foundation for future research into how model training strategies can impact explanation plausibility.

Our study's scope was constrained to a single data source (Beth Israel Deaconess Medical Center's MIMIC-III and MDACE) and one model architecture (PLM-ICD). The effectiveness of Attention and AttInGrad may vary with different architectures, particularly those employing multi-head attention and skip connections in the final layer. Further research is needed to investigate the generalizability of our findings across diverse model architectures, medical coding systems, languages, and healthcare institutions.

Furthermore, the limited size of the MDACE test set constrained our study, resulting in low statistical power for many experiments. Despite the desire to conduct more trials with various seeds, we limited ourselves to ten seeds per training strategy due to the high computational costs involved. Conducting more experiments or expanding the test set might have revealed nuances and differences that our initial setup failed to detect. Nevertheless, our results across runs, explanation methods, and analysis point in the same direction. Moreover, while the test set in the main paper only comprises 61 examples, each example contains 14 medical codes, each annotated with multiple evidence spans, providing greater statistical power. Finally, our comparison of the unsupervised approaches on the larger test set in \Cref{app:unsupervised} demonstrated similar results as on the smaller test set in the main paper. We, therefore, believe that our claims in this paper are well substantiated with empirical evidence.

\section*{Ethics statement}
Healthcare costs are continuously increasing worldwide, with administrative costs being a significant contributing factor~\cite{tsengAdministrativeCostsAssociated2018}. In this paper, we propose methods that may help reduce these administrative costs by making the review of medical code suggestions easier and faster. The aim of this paper was to develop technology to assist medical coders in performing tasks faster instead of replacing them.

Plausible but unfaithful explanations may risk convincing medical coders to accept medical code suggestions that are incorrect, thereby risking the patient's safety~\cite{jacoviFaithfullyInterpretableNLP2020}. We, therefore, advocate faithfulness to be of higher priority than in previous studies. 

Electronic healthcare records contain private information. The healthcare records in MIMIC-III, the dataset used in this paper, have been anonymized and stored in encrypted data storage accessible only to the main author, who has a license to the dataset and HIPAA training.

\section*{Acknowledgements}
This research was partially funded by the Innovation Fund Denmark via the Industrial Ph.D. Program (grant no. 2050-00040B) and Research Council of Finland (grant no. 322653). We thank Jonas Lyngsø for insightful discussions and code for dataloading. Furthermore, we thank Simon Flachs, Lana Krumm, and Andreas Geert Motzfeldt for revisions.

\bibliography{references}

\appendix
\section{Model architecture details}\label{app:model}
\begin{figure}[t]
    \centering
    \includegraphics[width=\linewidth]{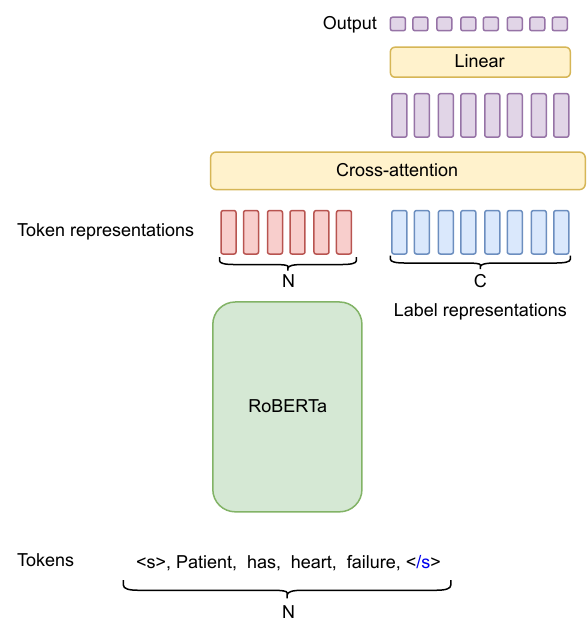}
    \caption{The PLM-CA architecture we used in our experiments.}
    \label{fig:PLM-Cross}
\end{figure}

PLM-ICD is a state-of-the-art automated medical coding model~\cite{huangPLMICDAutomaticICD2022, edinAutomatedMedicalCoding2023}. It comprises 131 million parameters. We experienced that PLM-ICD occasionally crashed during training. Therefore, we modified the architecture and called it pre-trained language model with class-wise cross attention (PLM-CA)~\cite{huangPLMICDAutomaticICD2022}.  Our architecture comprises an encoder and a decoder (see \Cref{fig:PLM-Cross}). The encoder transforms a sequence of tokens indices $\bm{t} \in \{0,1, \dots ,V\}^N$ into a sequence of contextualized token representations $\bm{H} \in \mathbb{R}^{N \times D}$. Both PLM-ICD and PLM-CA use RoBERTa-PM, a transformer pre-trained on PubMed articles and clinical notes, as the encoder~\cite{lewisPretrainedLanguageModels2020}. 

Our decoder takes the token representations $\bm{H}$ as input and outputs a sequence of output probabilities $\bm{\hat{y}} \in [0,1]^J$. It computes the output probabilities from the contextualized token representations using the following equation:
\begin{align}
    \bm{K} &= \bm{H}\bm{W}_{\text{key}} & \bm{V} &= \bm{H}\bm{W}_{\text{value}}
\end{align}
\begin{equation}
    \bm{A}_j = \text{softmax}(\bm{C}_j \bm{K}^T)
\label{eq:attention}
\end{equation}
\begin{equation}
    \bm{\hat{y}}_j = \text{sigmoid}(\text{layernorm}(\bm{A}_j \bm{V}) \bm{W}_{\text{out}})
\end{equation}

Where $\bm{W}_{\text{key}}\in \mathbb{R}^{D \times D}$, $\bm{W}_{\text{value}} \in \mathbb{R}^{D \times D}$, and $\bm{W}_{\text{out}} \in \mathbb{R}^{D}$ are learnable weights, $\bm{C} \in \mathbb{R}^{J \times D}$ is a sequence of learnable class representations, $\bm{A} \in \mathbb{R}^{J \times N}$ is the attention matrix, and $J$ is the number of classes.
In addition to being more stable during training, we also found that PLM-CA outperforms PLM-ICD on most metrics (see \Cref{tab:plm_prediction}). 

\begin{table}
    \centering
    \caption{Prediction performance on discharge summaries from the MIMIC-III full. The models are trained on the MIMIC-III full training set. The results are in percentages.}
    \label{tab:plm_prediction}
    \begin{tabular}{lccc}
\toprule

& \multicolumn{3}{c}{MIMIC-III full test} \\
\cmidrule(lr){2-4} 
Model & F1 micro & F1 macro & mAP \\
\midrule
PLM-ICD & 59.5$\pm$0.2 & 23.3$\pm$0.6 & 64.3$\pm$0.2\\
PLM-CA  & \textbf{60.0$\pm$0.1}  & \textbf{24.7$\pm$0.5} & \textbf{64.7$\pm$0.1}\\

\bottomrule
\end{tabular}
\end{table}

\section{Feature attribution methods}\label{sec:explanations}

\paragraph{Attention}
We use the raw attention weights $A_j$ (see \Cref{eq:attention}) in the cross-attention layer to explain class $j$. As mentioned in \Cref{seq:emc}, this explanation method was used by most previous studies in automated medical coding~\cite{mullenbachExplainablePredictionMedical2018, kimCanCurrentExplainability2022, dongExplainableAutomatedCoding2021, tengExplainablePredictionMedical2020, chengMDACEMIMICDocuments2023}. 

\paragraph{Attention Rollout (Rollout)}
The attention matrix in the cross-attention layer extracts information from the contextualized token representations encoded by RoBERTa (see \Cref{fig:PLM-Cross}). The token representations are not guaranteed to be aligned with the input tokens. A token representation at position $n$ could represent any and multiple tokens in the document. Attention rollout considers all the model's attention layers to calculate the feature attributions~\cite{abnarQuantifyingAttentionFlow2020a}. First, the attention matrices in each layer are averages across the heads. Then, the identity matrix is added to each layer's attention matrix to represent the skip connections. Finally, the attention rollout is calculated recursively using \Cref{eq:rollout}.

\begin{equation}
\bm{\tilde{A}}^{(l)} = 
\begin{cases} 
\bm{\bar{A}}^{(l)}\cdot\bm{\tilde{A}}^{(l-1)} & \text{if } l > 0 \\
\bm{\bar{A}}^{(l)} & \text{if } l = 0
\end{cases}
\label{eq:rollout}
\end{equation}

where $\bm{\tilde{A}} \in \mathbb{R}^{N \times N}$ is the rollout attention, and $\bm{\bar{A}} \in \mathbb{R}^{N \times N}$ is the attention averaged across heads with the added identity matrix. We calculated the final feature attribution score by multiplying the rollout attention from the final layer with the attention matrix from the cross-attention layer: $\bm{A} \cdot \bm{\tilde{A}}^{(L)}$, where $L$ is the number of attention layers.

\paragraph{Occlusion@1}
Occlusion@1 calculates each feature's score by occluding it and measuring the change in output confidence. The change of output will be the feature's score~\cite{,ribeiroWhyShouldTrust2016}. 

\paragraph{LIME}
Local Interpretable Model-agnostic Explanations (LIME) randomly occlude sets of tokens from a specific input and measure the change in output confidence. It uses these measurements to train a linear regression model that approximates the explained model's reasoning for that particular example. It then uses the linear regression weights to approximate each feature's influence~\cite{ribeiroWhyShouldTrust2016}.

\paragraph{KernelSHAP}
Shapley Additive Explanations (SHAP) is based on Shapley values from cooperative game theory, which fairly distributes the payout among players by considering each player's contribution in all possible coalitions. Ideally, SHAP quantifies all possible feature combinations in an input by occluding them and measuring the impact. However, this would result in $N\!$ forwards passes. KernelSHAP employs the LIME framework to approximate Shapley values using a weighted linear regression approach efficiently. We refer the reader to the seminal paper introducing SHAP and KernelSHAP for more details~\cite{lundbergUnifiedApproachInterpreting2017}.

\paragraph{InputXGrad}
InputXGradient multiplies the input gradients with the input~\cite{shrikumarLearningImportantFeatures2017}. We used the L2 norm to get the final feature attribution scores. We calculated the feature attribution scores for class $J$ as follows:

\begin{equation}
    \begin{bmatrix}
        \norm{\bm{X}_1 \odot \frac{\partial f_j}{\partial \bm{X}_1}(X)}_2
        \\
        \vdots
        \\
        \norm{\bm{X}_N \odot\frac{\partial f_j}{\partial \bm{X}_N}(X)}_2
    \end{bmatrix} 
\end{equation} 
\noindent where $\bm{X} \in \mathbb{R}^{N \times D}$ is the input token embeddings, $\odot$ is the element-wise matrix multiplication operation, $D$ is the embedding dimension, $N$ are the number of tokens in a document, and $J$ is the number of classes.

\paragraph{Integrated Gradients (IntGrad)}
Integrated Gradients (IntGrad) assigns an attribution score to each input feature by computing the integral of the gradients along the straight line path from the baseline $\bm{B}$ to the input $\bm{X}$~\cite{sundararajanAxiomaticAttributionDeep2017a}. Similar to InputXGradient, we used the L2-norm of the output to get the final attribution scores.

\paragraph{Deeplift}
DeepLIFT (Deep Learning Important FeaTures) backpropagates the contributions of all neurons in the model to every input feature~\cite{shrikumarLearningImportantFeatures2017}. It compares each neuron's activation to its baseline activation and assigns attribution scores according to the difference.

\paragraph{AttnGrad}
AttnGrad multiplies the attention Attention with the gradient of the model's output with respect to the attention weights ~\cite{serranoAttentionInterpretable2019}:

\begin{equation}
    \begin{bmatrix}
        \bm{A}_{j1} \cdot |\frac{\partial f_j}{\partial \bm{A}_{j1}}(\bm{X})|
        \\
        \vdots
        \\
        \bm{A}_{jN} \cdot|\frac{\partial f_j}{\partial \bm{A}_{jN}}(\bm{X})|
    \end{bmatrix} 
\end{equation} 
\noindent where $\bm{A} \in \mathbb{R}^{J \times N}$ is the attention matrix,  $N$ are the number of tokens in a document, and $J$ is the number of classes.

\paragraph{AttInGrad}
We found Attention Rollout to perform poorly on our task. Therefore, we developed a simple alternative approach to incorporate the impact of neighboring tokens into the attention explanations. AttInGrad incorporates the context by multiplying the attention $A_j$ with the InputXGrad feature attributions:

\begin{equation}
    \begin{bmatrix}
        \bm{A}_{j1} \cdot \norm{\bm{X}_1 \odot \frac{\partial f_j}{\partial \bm{X}_1}(X)}_2
        \\
        \vdots
        \\
        \bm{A}_{jN} \cdot\norm{\bm{X}_N \odot\frac{\partial f_j}{\partial \bm{X}_N}(X)}_2
    \end{bmatrix} 
\end{equation} 
\noindent where $\bm{A} \in \mathbb{R}^{J \times N}$ is the attention matrix, $\odot$ is the element-wise matrix multiplication operation, $N$ are the number of tokens in a document, and $J$ is the number of classes.

\section{Faithfulness evaluation metrics}\label{app:faithful}
Our faithfulness metrics, Sufficiency, and Comprehensiveness evaluate model output changes when important or unimportant features were masked~\cite{deyoungERASERBenchmarkEvaluate2020}.

\paragraph{Sufficiency} measures how masking non-important features affects the output. A high sufficiency score indicates that many low-attribution features significantly impact the model's output, suggesting the presence of false negatives. We calculated sufficiency using the following equation:

\begin{equation}
    \frac{1}{K}\sum_{i=N-K}^{N} \frac{\max(0, f(\bm{X})-f(\bm{R}_i))}{f(\bm{X})}
\end{equation}

Where $\bm{R}_i\in \mathbb{R}^{N \times D}$ 
represents the input with the $i$th least important feature replaced by mask tokens, $N$ is the number of tokens in an example, and $K$ is a hyperparameter.

\paragraph{Comprehensiveness} measures how masking important features affects the output. A high comprehensiveness score indicates that features with high attribution scores strongly influence the model's output, while a low score suggests many false positives. We calculated comprehensiveness using the following equation:

\begin{equation}
    \frac{1}{K}\sum_{i=0}^{K} \frac{\max(0, f(\bm{X})-f(\bm{\bar{R}}_i))}{f(\bm{X})}
\end{equation}

Where $\bm{\bar{R}}_i\in \mathbb{R}^{N \times D}$ denotes the input features with the $i$th highest attribution scores replaced by mask tokens.

We set $K=100$ because including all features led to sufficiency scores close to zero and comprehensiveness scores close to one, making it difficult to distinguish differences. Considering fewer features also made evaluation faster.

\section{Training details}\label{sec:hyperparameter}
We used the same hyperparameter as \citet{edinAutomatedMedicalCoding2023}. We trained for 20 epochs with the ADAMW optimizer~\cite{loshchilovDecoupledWeightDecay2022}, learning rate at $5\cdot 10^{-5}$, dropout at 0.2, no weight decay, and a linear decay learning rate scheduler with warmup.  We found the optimal hyperparameters for the auxiliary adversarial robustness training objectives through random search. For each training strategy, we searched the following options: 
learning rate: $\{5\cdot 10^{-5}$, $1\cdot 10^{-5}\}$, $\lambda_1$, $\lambda_2$, $\lambda_3$, $\beta$: $\{ 1.0,0.5,0.1,10^{-2},10^{-3},10^{-4},10^{-5}, 10^{-6}\}$, 
and $\epsilon$: $\{10^{-3},10^{-4},10^{-5}, 10^{-6}\}$. 
We found these hyperparameters to be optimal:
$\lambda_1=10^{-5}$, $\lambda_2=0.5$, $\lambda_3=0.5$, $\epsilon=10^{-5}$, and $\beta=0.01$. The learning rate was optimal at $5\cdot 10^{-5}$ for all training strategies except for token masking, where $1\cdot 10^{-5}$ was optimal. We optimized the token mask and adversarial noise using the ADAMW optimizer. In token-masking, we initialized the student and teacher model from a trained B$_{\text{U}}$. We fine-tuned the student for one epoch. We used the same hyperparameters as \citet{chengMDACEMIMICDocuments2023} for the supervised training strategy. 

We did not preprocess the text except to truncate the documents to a maximum length of 6000 tokens to reduce memory usage. Truncation is a common strategy in automated medical coding and has a negligible negative impact because few documents exceed the 6000 token limit ~\cite{edinAutomatedMedicalCoding2023}. 

\section{Additional results}
Because of space constraints, we could not include all of our results in the main paper. In this section, we present the excluded results:

\begin{enumerate}
    \item We show that the adversarial robustness training strategies do not affect the model's prediction performance.
    \item We present the results for all feature attribution methods, including Rand, AttGrad, Rollout, Occlusion@1, LIME, and KernelSHAP.
    \item We demonstrate that when the model struggles to predict the correct code, the explanations' plausibility drastically drops.
    \item We analyze if the robust models use fewer features by comparing the entropy of the feature attribution scores.
    \item We compare the unsupervised models on a bigger test set comprising 242 examples instead of 61.
\end{enumerate} 

\subsection{Advesarial training does not affect code prediction performance}
Previous papers have demonstrated that adversarial robustness often comes at the cost of accuracy~\cite{liFaithfulExplanationsText2023, tsiprasRobustnessMayBe2018}. Therefore, we evaluated whether the training strategies impacted the models' medical code prediction capabilities. As shown in \Cref{tab:prediction}, all models performed similarly on the MDACE test set. We also observed negligible performance differences on the MIMIC-III full test set. 

\begin{table}
    \centering
    \caption{Prediction performance on discharge summaries from the MDACE test sets. The results are in percentages. We use two classification metrics (F1 micro and macro) and mean Average Precision (mAP), a ranking metric.}
    \label{tab:prediction}
    \begin{tabular}{lccc}
\toprule

& \multicolumn{3}{c}{MDACE test} \\
\cmidrule(lr){2-4} 
Model & F1 micro $\uparrow$ & F1 macro $\uparrow$ & mAP $\uparrow$\\
\midrule
B$_{\text{S}}$  & 67.5$\pm$0.4  &51.5$\pm$1.1 & 72.0$\pm$0.7\\
B$_{\text{U}}$ & 67.7$\pm$0.3 & 51.6$\pm$0.7 & 72.0$\pm$0.4\\
IGR  & 68.0$\pm$0.6  & 52.2$\pm$1.2 & 72.4$\pm$1.0\\
TM  & 68.1$\pm$0.6 & 51.8$\pm$1.5 & 72.3$\pm$0.4\\
PGD & 68.1$\pm$ 0.5 & 52.1$\pm$ 1.2 & 72.0$\pm$ 0.7\\

\bottomrule
\end{tabular}
\end{table}

\subsection{Results from all feature attribution methods}\label{sec:full_results}
In the main paper, we only presented the results of selected feature attribution methods because of space constraints. Here, we present the results for all the feature attribution methods: Attention, AttGrad, Attention Rollout (Rollout), InputXGrad, Integrated gradients (IntGrad), Deeplift, and AttInGrad. We compare these methods with a random baseline (Rand), which randomly generates attribution scores. We present the plausibility results in \Cref{tab:plausibility_both_sup_full}, and the faithfulness results in \Cref{tab:fatihfulness_sup}. 

We did not include Occlusion@1, LIME, and KernelSHAP in these tables because they were too slow to calculate. We used the Captum implementation of the algorithms~\cite{kokhlikyanCaptumUnifiedGeneric2020}. It took around 45 minutes on an A100 GPU to calculate the explanations for a single example with LIME and KernelSHAP. Therefore, we only evaluated these methods on a single trained instance of B$_{\text{U}}$. We present the results in \Cref{tab:perturbation}.

\subsection{Relationship between confidence scores and explanation plausibility}
\label{sec:rel_plaus}
In \Cref{tab:plausibility_tp} and \Cref{tab:plausibility_fn}, we investigate the difference in explanation plausibility when the model correctly predicts an annotated code (true positive) and when it fails to predict an annotated code (false negative). The explanations are substantially better when the model correctly predicts the codes. 

\subsection{Entropy of explanation methods}\label{sec:entropy}
We calculated the entropy of the feature attribution distributions to test our hypothesis that robust training strategies reduce the number of features the model uses (see \Cref{tab:entropy}). The training strategies did not reduce the entropy. While we would expect a reduced entropy if the model used fewer features, other feature attribution distribution differences may simultaneously increase the entropy. The analysis is, therefore, inconclusive.

\begin{table}[t]
    \centering
    \caption{Entropy of the explanations.}
    \label{tab:entropy}
    \begin{tabular}{ccc}
    \toprule
        Explanation & Model & Entropy $\downarrow$\\
        \midrule
        \multirow[c]{2}{*}{InputXGrad} & B$_{\text{U}}$ & 0.74$\pm$0.01\\
        & IGR & 0.74$\pm$0.00\\
        & TM & 0.73$\pm$0.00\\
        \midrule
        \multirow[c]{2}{*}{Attention} & B$_{\text{U}}$ & 0.50$\pm$0.01 \\
        & IGR & 0.50$\pm$0.01\\
        & TM & 0.49$\pm$0.02\\
        \midrule
        \multirow[c]{2}{*}{AttInGrad} & B$_{\text{U}}$ & 0.28$\pm$0.02\\
        & IGR & 0.28$\pm$0.02\\
        & TM & 0.28$\pm$0.02\\
        \bottomrule
    \end{tabular}
    
\end{table}

\subsection{Unsupervised comparison on bigger test set}\label{app:unsupervised}
We included additional experiments on the unsupervised training strategies on a bigger test set. Since only the supervised training strategy required evidence-span annotations in the training set, we retrained our unsupervised methods on the MIMIC-III full training set and evaluated them on the MDACE training and test set (242 examples). 

\begin{figure*}
    \centering 
    \includegraphics[width=0.5\linewidth]{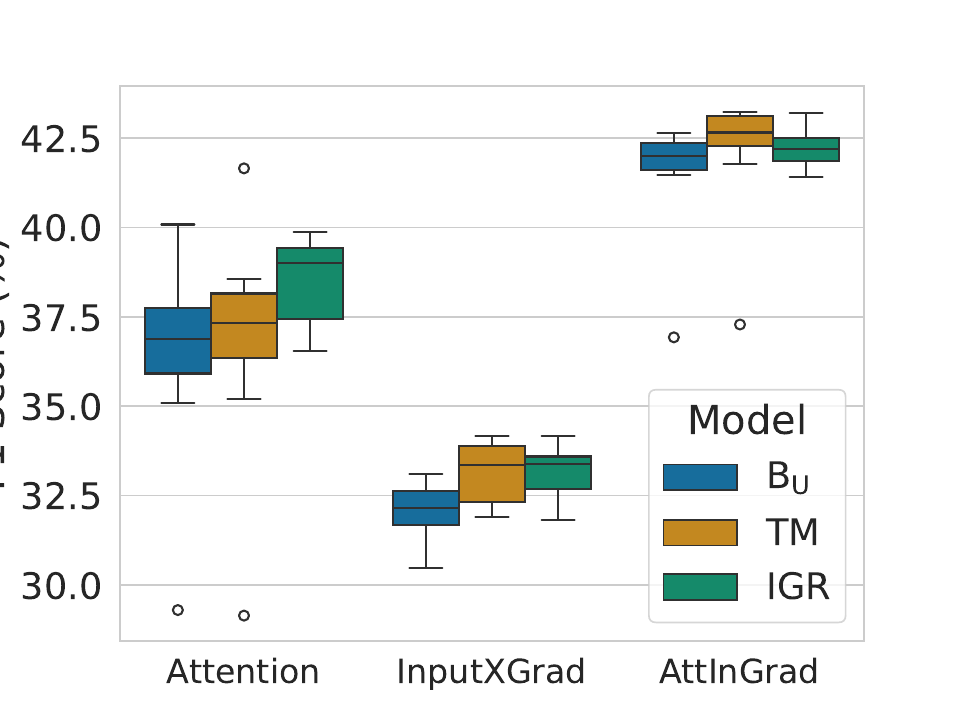}
    \includegraphics[width=0.5\linewidth]{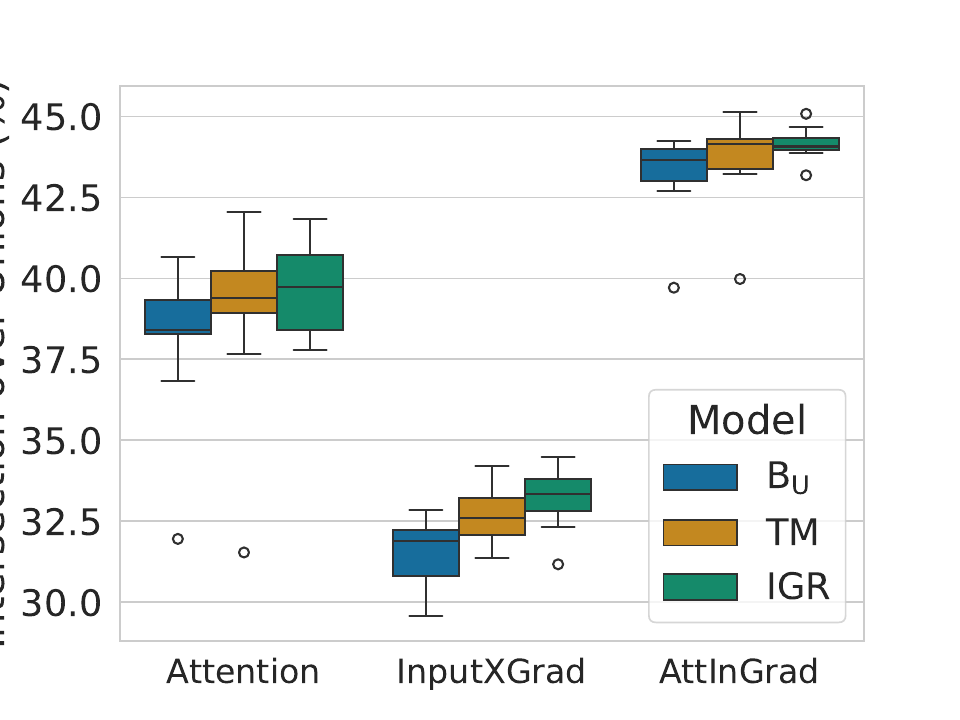}
    \caption{F1 scores and IOU of the unsupervised approaches on the bigger test set. Notice that there are no outliers for the attention-based explanations produced by IGR, which explains its higher mean scores in \Cref{tab:plausibility_both_uns}.}
    \label{fig:f1_unsupervised}
\end{figure*}

We present the plausibility results in \Cref{tab:plausibility_both_uns}. We observe that the results are similar to those of the main paper. However, the IGR produced substantially better attention-based explanations than in the main paper. In \Cref{fig:f1_unsupervised}, we inspect the inter-seed variance. We observe that IGR has no outliers. We, therefore, attribute the differences between this comparison and that in the main paper to none of the ten IGR runs happening to produce an outlier model. These results highlight the fragility of evaluating the attention-based feature attribution methods.

\section{CO$_2$ emissions}
Experiments were conducted using a private infrastructure, which has a carbon efficiency of 0.185 kgCO$_2$eq/kWh. To train B$_{\text{U}}$ or B$_{\text{S}}$, a cumulative of 8 hours of computation was performed on hardware of type A100 PCIe 40/80GB (TDP of 250W). Total emissions for one run are estimated to be 0.37 kgCO$_2$eq. The adversarial robustness training strategies required more hours of computation, therefore causing higher emissions. Input gradient regularization and projected gradient regularization required approximately 36 hours each (1.67 kgCO$_2$eq), while token masking required 2.5 hours of fine-tuning of B$_{\text{U}}$ (0.09 kgCO$_2$eq). We ran each experiment 10 times, resulting in total emissions of 41.86 kgCO$_2$eq, which is equivalent to burning 20.9 Kg of coal. Estimations were conducted using the \href{https://mlco2.github.io/impact#compute}{MachineLearning Impact calculator}~\cite{lacosteQuantifyingCarbonEmissions2019}. 

\section{Licenses}
We used MIMIC-III version 1.4, which is distributed under a non-commercial license, as detailed here: \url{https://physionet.org/content/mimiciii/view-license/1.4/}. Consequently, all model weights released in this paper are also restricted to non-commercial use. However, the MDACE annotations and our code are available under the MIT License. We provide instructions to how to obtain the datasets in our GitHub repository.

\begin{table*}
\centering

\caption{Plausibility comparison of Attention, AttGrad, Attention Rollout (Rollout), InputXGrad, Integrated gradients (IntGrad), Deeplift, and AttInGrad on the MDACE test set. Rand randomly generated attribution scores. Each experiment was run with ten different seeds. We show the mean of the seeds $\pm$ as the standard deviation. All the scores are presented as percentages. Bold numbers outperform the unsupervised baseline model, while underlined numbers outperform the supervised model.}
\label{tab:plausibility_both_sup_full}
\resizebox{\textwidth}{!}{%
\begin{tabular}{llcccccccccccc}
\toprule
 &  & \multicolumn{7}{c}{Prediction} & \multicolumn{3}{c}{Ranking}  \\
 \cmidrule(lr){3-9} \cmidrule(lr){10-12}
 Explainer& Model & P $\uparrow$ & R $\uparrow$ & F1 $\uparrow$ & AUPRC $\uparrow$ & Empty $\downarrow$ & SpanR $\uparrow$ & Cover $\uparrow$  & IOU $\uparrow$ & P@5 $\uparrow$ & R@5 $\uparrow$  \\ 
\midrule
\multirow[c]{4}{*}{Rand} & B$_{\text{S}}$ & 0.3$\pm$0.1 & 1.3$\pm$0.8 & 0.4$\pm$0.0 & 0.0$\pm$0.0 & 49.2$\pm$37.6 & 1.6$\pm$0.9 & 3.2$\pm$1.8 & 0.1$\pm$0.0 & 0.1$\pm$0.0 & 0.1$\pm$0.1 \\
 & B$_{\text{U}}$ & 0.4$\pm$0.1 & 1.0$\pm$0.7 & 0.4$\pm$0.1 & 0.0$\pm$0.0 & 64.7$\pm$37.9 & 1.1$\pm$0.7 & 2.3$\pm$1.6 & 0.1$\pm$0.1 & 0.1$\pm$0.1 & 0.2$\pm$0.1 \\
 & IGR & 0.4$\pm$0.2 & 0.7$\pm$0.7 & 0.4$\pm$0.2 & 0.0$\pm$0.0 & 80.3$\pm$30.9 & 1.0$\pm$0.9 & 1.9$\pm$1.9 & 0.1$\pm$0.0 & 0.1$\pm$0.1 & 0.2$\pm$0.1 \\
 & TM & 0.4$\pm$0.1 & 1.0$\pm$0.8 & 0.4$\pm$0.1 & 0.4$\pm$1.1 & 65.3$\pm$37.2 & 1.2$\pm$0.9 & 2.4$\pm$1.8 & 0.1$\pm$0.1 & 0.1$\pm$0.1 & 0.1$\pm$0.1 \\
  & PGD & 0.4$\pm$0.2 & 0.8$\pm$0.6 & 0.4$\pm$0.1 & 0.0$\pm$0.0 & 76.1$\pm$33.2 & 1.1$\pm$0.7 & 2.0$\pm$1.7 & 0.1$\pm$0.1 & 0.1$\pm$0.1 & 0.2$\pm$0.1 \\
 \midrule
\multirow[c]{4}{*}{Attention} & B$_{\text{S}}$ & 41.7$\pm$5.0 & 42.9$\pm$3.2 & 42.2$\pm$3.7 & 37.3$\pm$3.9 & 13.5$\pm$1.9 & 60.5$\pm$3.7 & 72.3$\pm$3.5 & 46.1$\pm$4.5 & 32.3$\pm$2.6 & 45.3$\pm$3.7 \\
 & B$_{\text{U}}$ & 35.9$\pm$5.9 & 37.7$\pm$5.7 & 36.6$\pm$5.3 & 31.4$\pm$6.4 & 21.1$\pm$3.6 & 53.2$\pm$7.4 & 63.4$\pm$6.8 & 39.3$\pm$6.9 & 28.8$\pm$4.2 & 40.3$\pm$5.9 \\
 & IGR & 35.5$\pm$6.2 & \textbf{40.4$\pm$6.1} & \textbf{37.7$\pm$5.9} & \textbf{32.4$\pm$6.8} & \textbf{19.4$\pm$2.1} & \textbf{55.5$\pm$8.0} & \textbf{64.5$\pm$6.2} & \textbf{40.0$\pm$7.4} & \textbf{29.2$\pm$4.5} & \textbf{40.9$\pm$6.3} \\
 & TM & \textbf{36.5$\pm$5.9} & \textbf{37.8$\pm$6.1} & \textbf{37.0$\pm$5.5} & \textbf{31.7$\pm$6.7} & 21.7$\pm$3.3 & \textbf{53.3$\pm$7.7} & 63.3$\pm$6.9 & \textbf{40.1$\pm$7.3} & \textbf{29.1$\pm$4.4} & \textbf{40.8$\pm$6.2} \\
  & PGD & 33.0$\pm$8.4 & \textbf{38.4$\pm$8.0} & 35.4$\pm$8.3 & 30.0$\pm$9.1 & \textbf{19.1$\pm$1.6} & 51.5$\pm$13.6 & 61.1$\pm$9.9 & 35.9$\pm$10.8 & 27.7$\pm$6.3 & 38.9$\pm$8.8 \\
 \midrule
 \multirow[c]{4}{*}{Rollout} & B$_{\text{S}}$ & 1.6$\pm$0.1 & 23.4$\pm$5.4 & 2.9$\pm$0.1 & 0.3$\pm$0.0 & 0.0$\pm$0.0 & 25.0$\pm$5.1 & 32.2$\pm$5.0 & 0.4$\pm$0.1 & 0.5$\pm$0.1 & 0.7$\pm$0.1 \\
 & B$_{\text{U}}$ & 1.7$\pm$0.1 & 24.1$\pm$2.4 & 3.1$\pm$0.2 & 0.3$\pm$0.0 & 0.0$\pm$0.0 & 25.1$\pm$2.8 & 31.4$\pm$2.9 & 0.3$\pm$0.1 & 0.5$\pm$0.1 & 0.7$\pm$0.1 \\
 & IGR & 1.7$\pm$0.1 & 24.3$\pm$2.5 & 3.1$\pm$0.2 & 0.3$\pm$0.0 & 0.0$\pm$0.0 & 25.4$\pm$3.1 & 31.6$\pm$2.9 & 0.3$\pm$0.1 & 0.5$\pm$0.1 & 0.7$\pm$0.1 \\
 & TM & 1.7$\pm$0.1 & 25.3$\pm$2.3 & 3.2$\pm$0.2 & 0.3$\pm$0.0 & 0.0$\pm$0.0 & 26.3$\pm$2.7 & 32.4$\pm$2.7 & 0.3$\pm$0.1 & 0.5$\pm$0.1 & 0.7$\pm$0.1 \\
  & PGD & 1.7$\pm$0.0 & 24.8$\pm$0.6 & 3.2$\pm$0.1 & 0.3$\pm$0.0 & 0.0$\pm$0.0 & 26.0$\pm$0.7 & 32.6$\pm$0.8 & 0.3$\pm$0.0 & 0.5$\pm$0.0 & 0.6$\pm$0.0 \\
 \midrule
\multirow[c]{4}{*}{$a \nabla a$} & B$_{\text{S}}$ & 37.4$\pm$5.8 & 37.8$\pm$3.0 & 37.4$\pm$4.1 & 31.5$\pm$5.1 & 11.9$\pm$2.4 & 55.0$\pm$3.8 & 69.9$\pm$3.8 & 39.7$\pm$6.1 & 29.4$\pm$3.1 & 41.2$\pm$4.4 \\
 & B$_{\text{U}}$ & 34.4$\pm$7.5 & 36.2$\pm$3.9 & 34.8$\pm$5.0 & 29.0$\pm$5.7 & 16.0$\pm$2.8 & 51.8$\pm$4.8 & 64.7$\pm$4.3 & 36.2$\pm$7.3 & 27.4$\pm$4.0 & 38.4$\pm$5.6 \\
 & IGR & \textbf{35.1$\pm$7.0} & \underline{\textbf{38.2$\pm$4.1}} & \textbf{36.3$\pm$5.2} & \textbf{30.8$\pm$6.4} & \textbf{15.0$\pm$2.2} & \textbf{54.2$\pm$5.5} & \textbf{65.7$\pm$4.0} & \textbf{38.3$\pm$9.0} & \textbf{28.9$\pm$4.6} & \textbf{40.5$\pm$6.5} \\
 & TM & \textbf{34.6$\pm$6.9} & \textbf{36.5$\pm$4.2} & \textbf{35.2$\pm$5.2} & \textbf{29.1$\pm$5.9} & 16.1$\pm$2.7 & \textbf{52.2$\pm$5.0} & \textbf{65.0$\pm$4.4} & \textbf{36.3$\pm$7.2} & \textbf{27.8$\pm$4.3} & \textbf{39.0$\pm$6.1} \\
  & PGD & 32.7$\pm$8.0 & \textbf{37.9$\pm$4.4} & \textbf{34.9$\pm$6.9} & 29.1$\pm$7.9 & \textbf{14.3$\pm$3.0} & \textbf{52.8$\pm$7.6} & 64.7$\pm$5.4 & 35.7$\pm$10.8 & \textbf{28.4$\pm$5.6} & \textbf{39.8$\pm$7.8} \\
 \midrule
\multirow[c]{4}{*}{InputXGrad} & B$_{\text{S}}$ & 30.7$\pm$2.0 & 33.7$\pm$2.6 & 32.0$\pm$1.4 & 24.9$\pm$2.0 & 7.3$\pm$1.3 & 48.5$\pm$2.8 & 64.0$\pm$2.8 & 32.2$\pm$2.1 & 26.6$\pm$1.2 & 37.3$\pm$1.6 \\
 & B$_{\text{U}}$ & 31.0$\pm$2.6 & 32.7$\pm$3.2 & 31.6$\pm$1.2 & 24.8$\pm$1.9 & 9.5$\pm$2.6 & 46.4$\pm$3.6 & 62.3$\pm$3.5 & 31.7$\pm$1.8 & 26.3$\pm$0.9 & 36.9$\pm$1.3 \\
 & IGR & \underline{\textbf{32.4$\pm$2.5}} & \underline{\textbf{33.8$\pm$2.2}} & \underline{\textbf{33.0$\pm$0.7}} & \underline{\textbf{27.0$\pm$0.7}} & 9.7$\pm$2.0 & \textbf{48.5$\pm$2.6} & \underline{\textbf{64.6$\pm$2.5}} & \underline{\textbf{32.8$\pm$1.0}} & \underline{\textbf{27.4$\pm$0.4}} & \underline{\textbf{38.4$\pm$0.5}} \\
 & TM & \underline{\textbf{32.4$\pm$2.7}} & \underline{\textbf{34.1$\pm$2.1}} & \underline{\textbf{33.1$\pm$1.4}} & \underline{\textbf{26.2$\pm$2.3}} & \textbf{8.6$\pm$1.6} & \textbf{48.5$\pm$2.4} & \underline{\textbf{64.8$\pm$2.5}} & \underline{\textbf{32.6$\pm$1.5}} & \underline{\textbf{27.4$\pm$1.1}} &\underline{\textbf{ 38.3$\pm$1.5}} \\
  & PGD & 30.1$\pm$2.2 & \textbf{32.9$\pm$1.7} & 31.4$\pm$1.4 & 24.8$\pm$1.5 & \textbf{9.3$\pm$2.2} & \textbf{46.6$\pm$2.1} & \textbf{62.6$\pm$2.6} & 31.4$\pm$1.3 & 26.1$\pm$1.0 & 36.6$\pm$1.4 \\
 \midrule
\multirow[c]{4}{*}{IG} & B$_{\text{S}}$ & 30.9$\pm$3.2 & 33.9$\pm$2.0 & 32.2$\pm$1.7 & 26.2$\pm$2.6 & 4.8$\pm$2.0 & 50.8$\pm$2.4 & 65.9$\pm$3.3 & 34.2$\pm$2.8 & 27.4$\pm$1.6 & 38.4$\pm$2.2 \\
 & B$_{\text{U}}$ & 31.3$\pm$4.3 & 32.8$\pm$2.7 & 31.9$\pm$3.2 & 26.0$\pm$4.3 & 5.2$\pm$1.0 & 49.4$\pm$3.0 & 64.4$\pm$3.1 & 33.9$\pm$4.2 & 26.5$\pm$2.5 & 37.1$\pm$3.6 \\
 & IGR & \underline{\textbf{32.6$\pm$2.2}} & \textbf{33.8$\pm$2.3} & \underline{\textbf{33.1$\pm$1.3}} & \underline{\textbf{26.6$\pm$2.0}} & 5.2$\pm$1.7 & \textbf{49.7$\pm$2.5} & \textbf{64.7$\pm$3.3} & \underline{\textbf{34.9$\pm$2.3}} & \underline{\textbf{27.6$\pm$0.9}} & \underline{\textbf{38.7$\pm$1.2}} \\
 & TM & \underline{\textbf{33.1$\pm$3.6}} & \underline{\textbf{34.0$\pm$3.7}}& \underline{\textbf{33.4$\pm$2.8}} & \underline{\textbf{27.5$\pm$3.7}} & 5.4$\pm$1.6 & \underline{\textbf{51.0$\pm$3.9}} & \textbf{65.5$\pm$4.1} & \underline{\textbf{35.3$\pm$4.2}} & \underline{\textbf{27.6$\pm$2.1}} & \underline{\textbf{38.7$\pm$3.0}} \\
  & PGD & 30.0$\pm$5.4 & \textbf{33.5$\pm$3.3} & 31.4$\pm$4.2 & 25.4$\pm$4.8 & \textbf{5.1$\pm$1.8} & \textbf{49.7$\pm$3.6} & \textbf{64.5$\pm$3.9} & 33.6$\pm$4.8 & 26.4$\pm$3.0 & 36.9$\pm$4.2 \\
 \midrule
\multirow[c]{4}{*}{Deeplift} & B$_{\text{S}}$ & 29.2$\pm$2.3 & 34.2$\pm$3.0 & 31.3$\pm$1.2 & 24.1$\pm$1.8 & 6.4$\pm$1.8 & 48.9$\pm$3.4 & 65.2$\pm$3.3 & 31.1$\pm$1.9 & 26.2$\pm$1.1 & 36.7$\pm$1.5 \\
 & B$_{\text{U}}$ & 31.2$\pm$1.9 & 31.4$\pm$2.4 & 31.2$\pm$1.5 & 24.1$\pm$2.1 & 9.1$\pm$1.6 & 45.1$\pm$2.5 & 61.7$\pm$2.9 & 30.9$\pm$1.6 & 25.8$\pm$1.1 & 36.1$\pm$1.5 \\
 & IGR & \underline{\textbf{31.0$\pm$1.8}} & \textbf{33.7$\pm$1.4} & \underline{\textbf{32.2$\pm$0.6}} & \underline{\textbf{25.9$\pm$0.9}} & \textbf{8.6$\pm$1.0} & \textbf{48.3$\pm$1.5} & \textbf{64.8$\pm$1.7} & \underline{\textbf{31.6$\pm$1.0}} & \underline{\textbf{26.7$\pm$0.5}} & \underline{\textbf{37.4$\pm$0.7}} \\
 & TM & \underline{\textbf{30.2$\pm$2.4}} & \underline{\textbf{34.8$\pm$1.4}} & \underline{\textbf{32.2$\pm$1.4}} & \underline{\textbf{25.1$\pm$2.5}} & \textbf{6.8$\pm$1.5} & \underline{\textbf{49.1$\pm$1.8}} & \underline{\textbf{65.7$\pm$1.6}} & \underline{\textbf{31.4$\pm$1.8}} & \underline{\textbf{26.6$\pm$1.1}} & \underline{\textbf{37.3$\pm$1.6}} \\
  & PGD & 29.4$\pm$2.8 & \textbf{32.8$\pm$1.5} & 30.9$\pm$1.5 & 23.9$\pm$1.5 & \textbf{8.1$\pm$1.8} & \textbf{46.7$\pm$1.}8 & \textbf{63.3$\pm$2.2} & 30.7$\pm$1.4 & 25.5$\pm$0.9 & 35.8$\pm$1.2 \\
 \midrule

\multirow[c]{4}{*}{AttInGrad} & B$_{\text{S}}$ & 40.6$\pm$2.2 & 45.9$\pm$3.1 & 43.0$\pm$1.9 & 38.8$\pm$2.1 & 1.2$\pm$0.5 & 63.9$\pm$2.9 & 79.0$\pm$1.9 & 45.7$\pm$2.6 & 33.3$\pm$1.4 & 46.6$\pm$1.9 \\
 & B$_{\text{U}}$ & 40.7$\pm$2.9 & 42.5$\pm$4.4 & 41.5$\pm$3.2 & 37.1$\pm$3.6 & 3.0$\pm$1.3 & 59.3$\pm$5.4 & 74.9$\pm$5.2 & 43.5$\pm$4.0 & 32.0$\pm$2.1 & 44.9$\pm$3.0 \\
 & IGR & 38.8$\pm$4.5 & \textbf{44.0$\pm$4.0} & 41.2$\pm$4.0 & \textbf{37.2$\pm$4.5} & \textbf{2.3$\pm$0.7} & \textbf{60.3$\pm$5.3} & 74.5$\pm$4.6 & 43.3$\pm$4.9 & 31.9$\pm$2.7 & 44.7$\pm$3.8 \\
 & TM & 40.2$\pm$3.0 & \textbf{43.9$\pm$4.8} & \textbf{41.9$\pm$3.4} & \textbf{37.8$\pm$3.9} & \textbf{2.8$\pm$1.6} & \textbf{60.5$\pm$5.8} & \textbf{75.9$\pm$5.3} & \textbf{44.0$\pm$3.9} & \textbf{32.5$\pm$2.3} & \textbf{45.5$\pm$3.2} \\
 & PGD & 37.1$\pm$8.2 & 42.5$\pm$5.7 & 39.3$\pm$7.1 & 34.7$\pm$8.6 & \textbf{2.3$\pm$0.9} & 57.7$\pm$9.1 & 72.7$\pm$7.4 & 39.6$\pm$10.2 & 30.9$\pm$4.8 & 43.3$\pm$6.7 \\
\bottomrule
\end{tabular}}
\end{table*}

\begin{table*}
    \centering
    \caption{Faithfulness comparison on the MDACE test set. Each experiment was run with ten different seeds. We show the mean of the seeds $\pm$ the standard deviation. The bold numbers represent the best score for each model and metric, while the underscore represents better than the Attention explanation method.}
    \label{tab:fatihfulness_sup}
    \begin{tabular}{llcc}
\toprule
 Model & Explainer & Comp $\uparrow$ & Suff $\downarrow$\\

\midrule
\multirow[c]{8}{*}{$\text{B}_{\text{U}}$} & Rand & 0.03$\pm$0.02 & 0.92$\pm$0.08 \\
 & Attention & 0.82$\pm$0.08 & 0.29$\pm$0.07 \\
 & InputXGrad & 0.75$\pm$0.01 & \underline{\textbf{0.17$\pm$0.03}} \\
 & IG & 0.74$\pm$0.02 & \underline{0.19$\pm$0.04}\\
 & Deeplift & 0.76$\pm$0.02 & \underline{0.18$\pm$0.03} \\
 & Rollout & 0.39$\pm$0.02 & 0.47$\pm$0.08 \\
 & $a \nabla a$ & 0.77$\pm$0.07 & \underline{0.28$\pm$0.06} \\
 & AttInGrad & \underline{\textbf{0.86$\pm$0.04}} & \underline{0.22$\pm$0.05} \\
 \midrule
\multirow[c]{8}{*}{$\text{B}_{\text{S}}$} & Rand & 0.02$\pm$0.02 & 0.93$\pm$0.08 \\
 & Attention & 0.84$\pm$0.06 & 0.42$\pm$0.15 \\
 & InputXGrad & 0.76$\pm$0.01 & \underline{\textbf{0.26$\pm$0.09}} \\
 & IG & 0.74$\pm$0.01 & \underline{0.28$\pm$0.10} \\
 & Deeplift & 0.76$\pm$0.01 & \underline{\textbf{0.26$\pm$0.09}} \\
 & Rollout & 0.39$\pm$0.02 & 0.47$\pm$0.08 \\
 & $a \nabla a$ & 0.77$\pm$0.06 & \underline{0.41$\pm$0.14} \\
 & AttInGrad & \underline{\textbf{0.86$\pm$0.03}} & \underline{0.34$\pm$0.12} \\
 \midrule
\multirow[c]{8}{*}{IGR} & Rand & 0.04$\pm$0.01 & 0.95$\pm$0.02 \\
 & Attention & 0.85$\pm$0.08 & 0.23$\pm$0.07 \\
 & InputXGrad & 0.77$\pm$0.01 & \underline{\textbf{0.14$\pm$0.03}} \\
 & IG & 0.75$\pm$0.02 & \underline{\textbf{0.14$\pm$0.03}} \\
 & Deeplift & 0.77$\pm$0.01 & \underline{\textbf{0.14$\pm$0.03}} \\
  & Rollout & 0.39$\pm$0.02 & 0.41$\pm$0.03 \\
 & $a \nabla a$ & 0.82$\pm$0.06 & 0.23$\pm$0.07 \\
 & AttInGrad & \underline{\textbf{0.87$\pm$0.04}} & \underline{0.17$\pm$0.05} \\
 \midrule
 \multirow[c]{8}{*}{TM} & Rand & 0.02$\pm$0.01 & 0.92$\pm$0.06 \\
 & Attention & 0.82$\pm$0.08 & 0.24$\pm$0.06 \\
 & InputXGrad & 0.76$\pm$0.01 & \underline{\textbf{0.15$\pm$0.03}} \\
 & IG & 0.74$\pm$0.01 & \underline{\textbf{0.15$\pm$0.03}} \\
 & Deeplift & 0.77$\pm$0.01 & \underline{\textbf{0.15$\pm$0.03}} \\
 & Rollout & 0.37$\pm$0.01 & 0.40$\pm$0.04 \\
 & $a \nabla a$ & 0.77$\pm$0.06 & 0.24$\pm$0.05 \\
 & AttInGrad & \underline{\textbf{0.86$\pm$0.04}} & \textbf{0.18$\pm$0.04} \\
 \midrule
\multirow[c]{8}{*}{PGD} & Rand & 0.04$\pm$0.01 & 0.93$\pm$0.06 \\
 & Attention & 0.84$\pm$0.10 & 0.28$\pm$0.08 \\
 & InputXGrad & 0.77$\pm$0.01 & \underline{\textbf{0.17$\pm$0.03}} \\
 & IG & 0.75$\pm$0.02 & \textbf{0.18$\pm$0.03} \\
 & Deeplift & 0.77$\pm$0.01 & \underline{\textbf{0.17$\pm$0.03}} \\
 & Rollout & 0.40$\pm$0.01 & 0.41$\pm$0.04 \\
 & $a \nabla a$ & 0.81$\pm$0.07 & 0.28$\pm$0.07 \\
 & AttInGrad & \underline{\textbf{0.86$\pm$0.05}} & \textbf{0.21$\pm$0.05} \\
\bottomrule
\end{tabular}
\end{table*}

\begin{table*}
\centering

\caption{Evaluation of perturbation-based feature attribution methods. We chose a random seed of B$_{\text{U}}$ and compared all feature attribution methods for that model. We have divided the feature attribution into attention-based, gradient-based, and perturbation-based. The highest values are bold, and the second highest are underlined.}
\label{tab:perturbation}
\resizebox{\textwidth}{!}{%
\begin{tabular}{lcccccccccccc}
\toprule
 & \multicolumn{7}{c}{Prediction} & \multicolumn{3}{c}{Ranking}  \\
 \cmidrule(lr){2-8} \cmidrule(lr){9-11}
 Explainer&  P $\uparrow$ & R $\uparrow$ & F1 $\uparrow$ & AUPRC $\uparrow$ & Empty $\downarrow$ & SpanR $\uparrow$ & Cover $\uparrow$  & IOU $\uparrow$ & P@5 $\uparrow$ & R@5 $\uparrow$  \\ 
\midrule
Attention & 37.7 & \underline{38.9} & \underline{38.3} & 33.1 & 14.8 & 55.1 & 72.4 & \underline{40.9} & \underline{29.7} & \underline{41.6} \\
$a \nabla a$ & \textbf{43.5} & 33.6 & 37.9 & 31.6 & 14.0 & 50.2 & 68.7 & 40.0 & 29.3 & 41.1 \\
AttInGrad & \underline{41.5} & 38.5 & \textbf{39.9} & \underline{35.7} & \underline{2.6} & \underline{55.3} & \underline{75.7} & \textbf{41.0} & \textbf{30.5} & \textbf{42.8} \\
\midrule
InputXGrad & 32.8 & 29.8 & 31.3 & 24.2 & 12.5 & 42.7 & 59.4 & 30.1 & 26.0 & 36.4 \\
IG & 33.5 & 35.4 & 34.4 & 29.3 & 5.3 & 51.6 & 66.1 & 38.2 & 27.7 & 38.9 \\
Deeplift & 30.0 & 32.1 & 31.0 & 23.6 & 9.6 & 44.8 & 61.6 & 28.7 & 25.1 & 35.1 \\
\midrule
Occl & 40.2 & 27.8 & 32.9 & 27.3 & 11.3 & 41.9 & 61.7 & 33.0 & 25.8 & 36.1 \\
KernelSHAP & 24.7 & 19.0 & 21.5 & \textbf{38.9} & 29.9 & 34.6 & 52.5 & 33.0 & 24.9 & 35.0 \\
LIME & 33.3 & \textbf{39.5} & 36.1 & 27.9 & \textbf{0.3} & \textbf{58.4} & \textbf{81.6} & 38.4 & 29.2 & 41.0 \\
\bottomrule
\end{tabular}}
\end{table*}

\begin{table*}
\centering
\caption{Plausibility scores. True positives}
\label{tab:plausibility_tp}
\resizebox{\textwidth}{!}{%
\begin{tabular}{llcccccccccccc}
\toprule
 &  & \multicolumn{7}{c}{Prediction} & \multicolumn{3}{c}{Ranking}  \\
 \cmidrule(lr){3-9} \cmidrule(lr){10-12}
 Explainer& Model & P $\uparrow$ & R $\uparrow$ & F1 $\uparrow$ & AUPRC $\uparrow$ & Empty $\downarrow$ & SpanR $\uparrow$ & Cover $\uparrow$  & IOU $\uparrow$ & P@5 $\uparrow$ & R@5 $\uparrow$  \\ 
 \midrule
\multirow[c]{5}{*}{Rand} & B$_{\text{S}}$ & 0.2$\pm$0.2 & 1.0$\pm$0.7 & 0.3$\pm$0.1 & 0.0$\pm$0.0 & 48.3$\pm$38.5 & 1.3$\pm$0.9 & 2.7$\pm$1.7 & 0.1$\pm$0.0 & 0.1$\pm$0.1 & 0.1$\pm$0.1 \\
 & B$_{\text{U}}$ & 0.3$\pm$0.2 & 0.7$\pm$0.6 & 0.3$\pm$0.1 & 0.0$\pm$0.0 & 64.1$\pm$38.6 & 0.9$\pm$0.7 & 1.8$\pm$1.5 & 0.1$\pm$0.1 & 0.2$\pm$0.1 & 0.2$\pm$0.1 \\
 & IGR & 0.3$\pm$0.2 & 0.6$\pm$0.7 & 0.3$\pm$0.1 & 0.0$\pm$0.0 & 79.9$\pm$31.7 & 0.8$\pm$0.9 & 1.5$\pm$1.9 & 0.1$\pm$0.0 & 0.1$\pm$0.1 & 0.2$\pm$0.1 \\
 & TM & 0.3$\pm$0.1 & 0.8$\pm$0.8 & 0.3$\pm$0.1 & 0.4$\pm$1.2 & 64.8$\pm$38.1 & 1.1$\pm$1.0 & 2.1$\pm$1.9 & 0.1$\pm$0.1 & 0.1$\pm$0.1 & 0.1$\pm$0.1 \\
 & PGD & 0.3$\pm$0.2 & 0.6$\pm$0.6 & 0.3$\pm$0.1 & 0.0$\pm$0.0 & 75.3$\pm$34.0 & 0.9$\pm$0.7 & 1.7$\pm$1.5 & 0.1$\pm$0.1 & 0.1$\pm$0.1 & 0.2$\pm$0.1 \\
\midrule \multirow[c]{5}{*}{Attention} & B$_{\text{S}}$ & 40.7$\pm$5.0 & 51.1$\pm$3.7 & 45.2$\pm$4.3 & 41.2$\pm$4.5 & 4.2$\pm$1.4 & 69.5$\pm$4.9 & 81.8$\pm$4.3 & 48.4$\pm$5.2 & 34.3$\pm$3.2 & 49.0$\pm$4.3 \\
 & B$_{\text{U}}$ & 35.1$\pm$6.0 & 45.8$\pm$7.0 & 39.6$\pm$6.0 & 35.5$\pm$7.5 & 9.1$\pm$3.3 & 62.7$\pm$9.1 & 73.5$\pm$7.7 & 43.3$\pm$8.2 & 31.0$\pm$4.9 & 44.5$\pm$7.1 \\
 & IGR & \underline{\textbf{35.3$\pm$6.5}} & \underline{\textbf{49.5$\pm$7.7}} & \underline{\textbf{41.1$\pm$6.8}} & \underline{\textbf{36.9$\pm$8.0}} & \textbf{7.5$\pm$1.6} & \underline{\textbf{65.8$\pm$10.0}} & \underline{\textbf{75.1$\pm$7.5}} & \underline{\textbf{44.2$\pm$8.3}} & \underline{\textbf{31.7$\pm$5.2}} & \underline{\textbf{45.4$\pm$7.3}} \\
 & TM & \underline{\textbf{35.7$\pm$6.0}} & \underline{\textbf{45.9$\pm$7.4}} & \underline{\textbf{40.0$\pm$6.2}} & \underline{\textbf{35.8$\pm$7.9}} & 9.7$\pm$3.4 & \underline{\textbf{62.9$\pm$9.6}} & \underline{\textbf{73.6$\pm$8.2}} & \underline{\textbf{44.0$\pm$8.5}} & \underline{\textbf{31.3$\pm$5.0}} & \underline{\textbf{44.8$\pm$7.2}} \\
 & PGD & 32.8$\pm$8.8 & \underline{\textbf{47.3$\pm$10.8}} & 38.7$\pm$9.8 & 34.5$\pm$10.9 & \textbf{7.3$\pm$1.2} & 61.6$\pm$16.9 & 71.8$\pm$12.9 & 40.4$\pm$12.3 & 30.1$\pm$7.1 & 43.1$\pm$10.1 \\
\midrule \multirow[c]{5}{*}{Rollout} & B$_{\text{S}}$ & 1.7$\pm$0.1 & 28.6$\pm$5.8 & 3.1$\pm$0.1 & 0.3$\pm$0.0 & 0.0$\pm$0.0 & 30.7$\pm$6.0 & 37.8$\pm$5.6 & 0.4$\pm$0.1 & 0.5$\pm$0.1 & 0.8$\pm$0.1 \\
 & B$_{\text{U}}$ & 1.8$\pm$0.1 & 28.6$\pm$2.6 & 3.3$\pm$0.1 & 0.3$\pm$0.0 & 0.0$\pm$0.0 & 29.8$\pm$3.1 & 35.7$\pm$3.5 & 0.4$\pm$0.0 & 0.5$\pm$0.1 & 0.8$\pm$0.1 \\
 & IGR & 1.8$\pm$0.1 & 29.3$\pm$2.9 & 3.4$\pm$0.2 & 0.3$\pm$0.0 & 0.0$\pm$0.0 & 30.7$\pm$3.6 & 36.9$\pm$3.4 & 0.4$\pm$0.0 & 0.5$\pm$0.1 & 0.8$\pm$0.1 \\
 & TM & 1.8$\pm$0.1 & 29.6$\pm$2.3 & 3.3$\pm$0.1 & 0.3$\pm$0.0 & 0.0$\pm$0.0 & 31.0$\pm$3.0 & 36.7$\pm$3.3 & 0.4$\pm$0.0 & 0.5$\pm$0.1 & 0.7$\pm$0.1 \\
 & PGD & 1.8$\pm$0.0 & 30.4$\pm$0.7 & 3.4$\pm$0.1 & 0.3$\pm$0.0 & 0.0$\pm$0.0 & 31.7$\pm$0.9 & 38.4$\pm$1.1 & 0.4$\pm$0.0 & 0.5$\pm$0.1 & 0.8$\pm$0.1 \\
\midrule \multirow[c]{5}{*}{$a \nabla a$} & B$_{\text{S}}$ & 35.9$\pm$5.7 & 43.1$\pm$3.2 & 39.0$\pm$4.4 & 33.7$\pm$5.3 & 4.4$\pm$1.8 & 60.8$\pm$4.0 & 76.5$\pm$4.3 & 40.7$\pm$6.5 & 30.8$\pm$3.5 & 43.9$\pm$4.8 \\
 & B$_{\text{U}}$ & 33.1$\pm$7.0 & 42.2$\pm$4.1 & 36.7$\pm$5.2 & 31.5$\pm$6.1 & 6.0$\pm$1.8 & 59.2$\pm$4.9 & 72.6$\pm$3.7 & 38.3$\pm$7.7 & 28.8$\pm$4.2 & 41.3$\pm$6.0 \\
 & IGR & \underline{\textbf{34.3$\pm$6.9}} & \underline{\textbf{45.0$\pm$4.5}} & \underline{\textbf{38.6$\pm$5.7}} & \underline{\textbf{33.7$\pm$7.0}} & \textbf{5.0$\pm$1.7} & \underline{\textbf{62.7$\pm$5.9}} & \underline{\textbf{74.4$\pm$3.9}} & \underline{\textbf{41.1$\pm$9.7}} & \underline{\textbf{30.6$\pm$4.7}} & \underline{\textbf{43.9$\pm$6.7}} \\
 & TM & \underline{\textbf{33.4$\pm$6.4}} & \underline{\textbf{42.6$\pm$4.8}} & \underline{\textbf{37.1$\pm$5.3}} & \underline{\textbf{31.6$\pm$6.3}} & \textbf{5.8$\pm$2.2} & \underline{\textbf{59.7$\pm$5.5}} & \underline{\textbf{73.1$\pm$4.3}} & \underline{\textbf{38.4$\pm$7.4}} & \underline{\textbf{29.1$\pm$4.5}} & \underline{\textbf{41.7$\pm$6.4}} \\
 & PGD & 32.2$\pm$8.0 & \underline{\textbf{44.9$\pm$5.7}} & \underline{\textbf{37.2$\pm$7.8}} & \underline{\textbf{32.1$\pm$8.9}} & \textbf{4.8$\pm$1.8} & \underline{\textbf{60.7$\pm$9.2}} & \underline{\textbf{73.3$\pm$7.0}} & \underline{\textbf{38.4$\pm$11.7}} & \underline{\textbf{30.0$\pm$5.8}} & \underline{\textbf{42.9$\pm$8.2}} \\
\midrule \multirow[c]{5}{*}{InputXGrad} & B$_{\text{S}}$ & 30.4$\pm$2.0 & 37.7$\pm$2.6 & 33.6$\pm$1.4 & 26.7$\pm$2.3 & 3.4$\pm$1.1 & 53.5$\pm$2.8 & 69.2$\pm$2.7 & 34.2$\pm$2.3 & 28.3$\pm$1.3 & 40.4$\pm$1.6 \\
 & B$_{\text{U}}$ & 30.9$\pm$2.6 & 37.0$\pm$3.8 & 33.4$\pm$1.2 & 26.8$\pm$2.1 & 4.6$\pm$1.9 & 51.9$\pm$4.2 & 67.6$\pm$3.6 & 34.8$\pm$1.9 & 28.1$\pm$1.0 & 40.3$\pm$1.6 \\
 & IGR & \underline{\textbf{32.0$\pm$2.7}} & \underline{\textbf{38.3$\pm$2.5}} & \underline{\textbf{34.7$\pm$1.0}} & \underline{\textbf{29.3$\pm$1.0}} & 4.9$\pm$1.5 & \underline{\textbf{54.4$\pm$2.8}} & \underline{\textbf{70.3$\pm$2.7}} & \underline{\textbf{35.8$\pm$1.5}} & \underline{\textbf{29.1$\pm$0.5}} & \underline{\textbf{41.7$\pm$0.6}} \\
 & TM & \underline{\textbf{32.5$\pm$2.3}} & \underline{\textbf{39.2$\pm$2.7}} & \underline{\textbf{35.4$\pm$1.3}} & \underline{\textbf{28.8$\pm$2.2}} & \textbf{3.7$\pm$1.3} & \underline{\textbf{55.0$\pm$3.0}} & \underline{\textbf{71.3$\pm$2.7}} & \underline{\textbf{35.5$\pm$1.5}} & \underline{\textbf{29.4$\pm$1.0}} & \underline{\textbf{42.2$\pm$1.4}} \\
 & PGD & 29.8$\pm$2.3 & \underline{\textbf{37.2$\pm$1.7}} & 33.0$\pm$1.7 & \underline{\textbf{27.1$\pm$1.7}} & 4.7$\pm$1.5 & \underline{\textbf{52.3$\pm$2.3}} & \underline{\textbf{68.0$\pm$2.9}} & 34.1$\pm$1.4 & 28.0$\pm$1.3 & 40.0$\pm$1.7 \\
\midrule \multirow[c]{5}{*}{IG} & B$_{\text{S}}$ & 31.6$\pm$3.6 & 37.3$\pm$2.6 & 34.1$\pm$2.4 & 27.8$\pm$3.2 & 2.1$\pm$1.2 & 55.3$\pm$2.6 & 70.6$\pm$3.8 & 36.7$\pm$3.3 & 29.0$\pm$1.8 & 41.5$\pm$2.8 \\
 & B$_{\text{U}}$ & 31.5$\pm$4.5 & 36.5$\pm$3.2 & 33.7$\pm$3.7 & 27.7$\pm$4.9 & 1.3$\pm$0.5 & 54.9$\pm$3.2 & 69.7$\pm$3.6 & 37.0$\pm$5.0 & 28.2$\pm$2.9 & 40.4$\pm$3.9 \\
 & IGR & \underline{\textbf{32.6$\pm$2.2}} & \underline{\textbf{37.4$\pm$2.4}} & \underline{\textbf{34.7$\pm$1.4}} & 27.7$\pm$2.5 & \textbf{1.2$\pm$0.6} & 54.9$\pm$2.8 & 69.5$\pm$3.6 & \underline{\textbf{37.5$\pm$2.6}} & \underline{\textbf{29.1$\pm$0.9}} & \underline{\textbf{41.7$\pm$1.2}} \\
 & TM & \underline{\textbf{33.7$\pm$3.6}} & \underline{\textbf{38.5$\pm$4.4}} & \underline{\textbf{35.8$\pm$3.2}} & \underline{\textbf{29.8$\pm$4.6}} & \underline{1.3$\pm$0.9} & \underline{\textbf{57.0$\pm$4.9}} & \underline{\textbf{71.5$\pm$4.7}} & \underline{\textbf{38.8$\pm$5.2}} & \underline{\textbf{29.6$\pm$2.6}} & \underline{\textbf{42.5$\pm$3.5}} \\
 & PGD & 30.4$\pm$5.5 & \underline{\textbf{37.2$\pm$3.9}} & 33.2$\pm$4.6 & 27.0$\pm$5.4 & \underline{1.8$\pm$1.0} & \underline{\textbf{55.2$\pm$4.4}} & \underline{\textbf{69.8$\pm$4.6}} & 36.6$\pm$5.5 & 28.1$\pm$3.3 & 40.2$\pm$4.8 \\
\midrule \multirow[c]{5}{*}{Deeplift} & B$_{\text{S}}$ & 29.1$\pm$2.3 & 38.4$\pm$3.0 & 33.0$\pm$1.3 & 26.1$\pm$1.9 & 3.2$\pm$1.3 & 54.2$\pm$3.6 & 70.5$\pm$3.4 & 33.4$\pm$2.1 & 27.7$\pm$1.2 & 39.6$\pm$1.5 \\
 & B$_{\text{U}}$ & 31.2$\pm$1.9 & 35.7$\pm$2.7 & 33.2$\pm$1.5 & 26.2$\pm$2.3 & 4.0$\pm$1.4 & 50.7$\pm$2.9 & 67.3$\pm$3.0 & 33.9$\pm$1.8 & 27.6$\pm$1.1 & 39.6$\pm$1.6 \\
 & IGR & \underline{30.6$\pm$2.0} & \underline{\textbf{38.2$\pm$1.5}} & \underline{\textbf{33.9$\pm$0.9}} & \underline{\textbf{28.1$\pm$1.2}} & \textbf{3.8$\pm$0.8} & \underline{\textbf{54.3$\pm$1.7}} & \underline{\textbf{70.5$\pm$1.7}} & \underline{\textbf{34.2$\pm$1.4}} & \underline{\textbf{28.5$\pm$0.6}} & \underline{\textbf{40.8$\pm$0.8}} \\
 & TM & \underline{30.3$\pm$2.1} & \underline{\textbf{39.9$\pm$1.9}} & \underline{\textbf{34.4$\pm$1.3}} & \underline{\textbf{27.7$\pm$2.3}} & \textbf{2.3$\pm$0.5} & \underline{\textbf{55.8$\pm$2.2}} & \underline{\textbf{72.5$\pm$1.7}} & \underline{\textbf{34.3$\pm$2.0}} & \underline{\textbf{28.6$\pm$1.0}} & \underline{\textbf{41.0$\pm$1.5}} \\
 & PGD & \underline{29.3$\pm$3.0} & \underline{\textbf{37.2$\pm$1.8}} & 32.6$\pm$1.8 & 26.1$\pm$1.8 & \textbf{3.8$\pm$1.3} & \underline{\textbf{52.5$\pm$2.3}} & \underline{\textbf{69.0$\pm$2.7}} & \underline{33.5$\pm$1.5} & 27.4$\pm$1.1 & 39.2$\pm$1.6 \\
\midrule \multirow[c]{5}{*}{AttInGrad} & B$_{\text{S}}$ & 41.8$\pm$2.4 & 50.2$\pm$3.2 & 45.6$\pm$2.1 & 41.8$\pm$2.3 & 0.0$\pm$0.1 & 68.7$\pm$3.2 & 83.4$\pm$2.2 & 48.1$\pm$2.7 & 35.0$\pm$1.5 & 50.0$\pm$1.9 \\
 & B$_{\text{U}}$ & 41.3$\pm$3.3 & 47.2$\pm$5.3 & 43.9$\pm$3.9 & 40.3$\pm$4.4 & 0.2$\pm$0.4 & 65.3$\pm$6.4 & 80.4$\pm$5.8 & 46.4$\pm$4.9 & 33.7$\pm$2.6 & 48.4$\pm$3.8 \\
 & IGR & 39.7$\pm$4.9 & \underline{\textbf{49.1$\pm$4.7}} & 43.9$\pm$4.7 & \underline{\textbf{41.0$\pm$5.4}} & \textbf{0.1$\pm$0.1} & \underline{\textbf{66.9$\pm$6.1}} & 80.4$\pm$4.8 & \underline{\textbf{46.6$\pm$5.5}} & \underline{\textbf{33.8$\pm$3.2}} & 48.4$\pm$4.5 \\
 & TM & 40.9$\pm$3.5 & \underline{\textbf{48.5$\pm$6.1}} & \underline{\textbf{44.3$\pm$4.3}} & \underline{\textbf{41.3$\pm$4.9}} & 0.3$\pm$0.6 & \underline{\textbf{66.3$\pm$7.3}} & \underline{\textbf{81.3$\pm$6.3}} & \underline{\textbf{47.1$\pm$5.2}} & \underline{\textbf{34.2$\pm$2.8}} & \underline{\textbf{49.0$\pm$4.3}} \\
 & PGD & 37.9$\pm$9.3 & 47.1$\pm$8.0 & 41.7$\pm$8.8 & 38.4$\pm$10.2 & 0.2$\pm$0.2 & 63.2$\pm$12.0 & 77.8$\pm$9.7 & 42.8$\pm$11.9 & 32.7$\pm$5.7 & 46.8$\pm$8.1 \\
\bottomrule
\end{tabular}
}

\end{table*}

\begin{table*}
\centering
\caption{Plausibility scores. False negatives}
\label{tab:plausibility_fn}
\resizebox{\textwidth}{!}{%
\begin{tabular}{llcccccccccccc}
\toprule
 &  & \multicolumn{7}{c}{Prediction} & \multicolumn{3}{c}{Ranking}  \\
 \cmidrule(lr){3-9} \cmidrule(lr){10-12}
 Explainer& Model & P $\uparrow$ & R $\uparrow$ & F1 $\uparrow$ & AUPRC $\uparrow$ & Empty $\downarrow$ & SpanR $\uparrow$ & Cover $\uparrow$  & IOU $\uparrow$ & P@5 $\uparrow$ & R@5 $\uparrow$  \\ 
 \midrule
\multirow[c]{5}{*}{Rand} & B$_{\text{S}}$ & 0.4$\pm$0.1 & 1.9$\pm$0.9 & 0.6$\pm$0.1 & 0.0$\pm$0.0 & 51.2$\pm$35.8 & 2.2$\pm$1.0 & 4.3$\pm$2.3 & 0.1$\pm$0.1 & 0.1$\pm$0.1 & 0.1$\pm$0.1 \\
 & B$_{\text{U}}$ & 0.4$\pm$0.2 & 1.4$\pm$0.9 & 0.6$\pm$0.2 & 0.0$\pm$0.0 & 65.8$\pm$36.6 & 1.6$\pm$1.0 & 3.4$\pm$1.9 & 0.0$\pm$0.1 & 0.1$\pm$0.1 & 0.1$\pm$0.1 \\
 & IGR & 0.5$\pm$0.2 & 1.1$\pm$0.9 & 0.6$\pm$0.3 & 0.0$\pm$0.0 & 81.2$\pm$29.0 & 1.4$\pm$1.1 & 2.6$\pm$2.1 & 0.0$\pm$0.0 & 0.1$\pm$0.1 & 0.1$\pm$0.1 \\
 & TM & 0.4$\pm$0.2 & 1.2$\pm$0.8 & 0.5$\pm$0.2 & 0.0$\pm$0.0 & 66.5$\pm$35.4 & 1.4$\pm$0.8 & 2.9$\pm$1.6 & 0.1$\pm$0.1 & 0.1$\pm$0.2 & 0.2$\pm$0.2 \\
 & PGD & 0.5$\pm$0.2 & 1.2$\pm$0.8 & 0.6$\pm$0.1 & 0.0$\pm$0.0 & 77.9$\pm$31.4 & 1.4$\pm$0.8 & 2.8$\pm$1.9 & 0.1$\pm$0.1 & 0.1$\pm$0.1 & 0.1$\pm$0.1 \\
\midrule \multirow[c]{5}{*}{Attention} & B$_{\text{S}}$ & 46.3$\pm$4.8 & 26.4$\pm$2.8 & 33.5$\pm$2.9 & 30.1$\pm$3.1 & 33.4$\pm$3.6 & 41.3$\pm$2.8 & 52.1$\pm$3.5 & 41.2$\pm$3.3 & 28.1$\pm$1.6 & 37.9$\pm$2.6 \\
 & B$_{\text{U}}$ & 40.1$\pm$6.0 & 21.4$\pm$3.1 & 27.7$\pm$3.6 & 23.8$\pm$4.0 & 47.2$\pm$5.4 & 32.5$\pm$4.0 & 41.6$\pm$5.3 & 30.6$\pm$4.3 & 23.8$\pm$2.8 & 31.9$\pm$3.7 \\
 & IGR & 36.6$\pm$5.1 & 21.1$\pm$2.7 & 26.8$\pm$3.3 & 23.3$\pm$4.2 & \textbf{46.0$\pm$3.1} & 32.0$\pm$3.6 & 40.4$\pm$3.5 & 30.4$\pm$5.4 & 23.6$\pm$3.1 & 31.5$\pm$4.2 \\
 & TM & \underline{\textbf{40.5$\pm$5.5}} & \underline{\textbf{21.7$\pm$4.0}} & \underline{\textbf{28.1$\pm$4.3}} & \underline{\textbf{24.0$\pm$4.7}} & 47.7$\pm$4.8 & \underline{\textbf{32.6$\pm$4.3}} & 41.2$\pm$5.3 & \underline{\textbf{31.9$\pm$5.1}} & \underline{\textbf{24.6$\pm$3.4}} & \underline{\textbf{32.9$\pm$4.6}} \\
 & PGD & 33.8$\pm$7.0 & 19.8$\pm$2.5 & 24.8$\pm$3.9 & 20.8$\pm$5.2 & \textbf{45.4$\pm$3.6} & 29.0$\pm$6.3 & 37.3$\pm$3.7 & 26.1$\pm$7.5 & 22.4$\pm$4.5 & 30.1$\pm$6.2 \\
\midrule \multirow[c]{5}{*}{Rollout} & B$_{\text{S}}$ & 1.2$\pm$0.2 & 12.9$\pm$4.7 & 2.2$\pm$0.3 & 0.2$\pm$0.0 & 0.0$\pm$0.0 & 13.0$\pm$3.5 & 20.1$\pm$4.2 & 0.3$\pm$0.3 & 0.4$\pm$0.1 & 0.5$\pm$0.2 \\
 & B$_{\text{U}}$ & 1.4$\pm$0.2 & 15.1$\pm$2.4 & 2.6$\pm$0.4 & 0.2$\pm$0.1 & 0.0$\pm$0.0 & 15.1$\pm$2.1 & 22.1$\pm$1.9 & 0.2$\pm$0.2 & 0.4$\pm$0.1 & 0.6$\pm$0.2 \\
 & IGR & 1.3$\pm$0.2 & 13.6$\pm$3.4 & 2.4$\pm$0.5 & 0.2$\pm$0.0 & 0.0$\pm$0.0 & 13.4$\pm$3.2 & 19.7$\pm$4.1 & 0.1$\pm$0.2 & 0.3$\pm$0.1 & 0.4$\pm$0.1 \\
 & TM & 1.5$\pm$0.2 & 16.6$\pm$2.6 & 2.7$\pm$0.4 & 0.2$\pm$0.1 & 0.0$\pm$0.0 & 16.0$\pm$2.3 & 23.0$\pm$2.3 & 0.2$\pm$0.2 & 0.4$\pm$0.1 & 0.5$\pm$0.1 \\
 & PGD & 1.3$\pm$0.1 & 13.1$\pm$0.9 & 2.3$\pm$0.2 & 0.2$\pm$0.0 & 0.0$\pm$0.0 & 13.4$\pm$1.4 & 19.7$\pm$2.4 & 0.1$\pm$0.0 & 0.3$\pm$0.0 & 0.3$\pm$0.1 \\
\midrule \multirow[c]{5}{*}{$a \nabla a$} & B$_{\text{S}}$ & 43.0$\pm$6.5 & 27.2$\pm$3.1 & 33.2$\pm$3.8 & 28.2$\pm$4.8 & 28.0$\pm$3.8 & 42.7$\pm$4.0 & 55.9$\pm$3.3 & 37.5$\pm$5.5 & 26.5$\pm$2.5 & 35.7$\pm$3.7 \\
 & B$_{\text{U}}$ & 39.7$\pm$9.5 & 24.1$\pm$4.0 & 29.6$\pm$5.1 & 24.8$\pm$5.4 & 37.5$\pm$5.4 & 35.8$\pm$5.3 & 47.5$\pm$6.1 & 31.5$\pm$6.7 & 24.2$\pm$4.0 & 32.4$\pm$5.4 \\
 & IGR & 39.0$\pm$7.8 & 23.8$\pm$3.5 & 29.2$\pm$4.3 & \underline{\textbf{25.8$\pm$5.4}} & 37.5$\pm$3.8 & 35.2$\pm$4.6 & 46.1$\pm$4.5 & \underline{\textbf{32.1$\pm$7.9}} & \underline{\textbf{25.0$\pm$4.5}} & \underline{\textbf{33.4$\pm$6.1}} \\
 & TM & \underline{\textbf{39.8$\pm$9.2}} & \underline{\textbf{24.3$\pm$4.2}} & \underline{\textbf{29.9$\pm$5.5}} & \underline{\textbf{25.2$\pm$5.8}} & 38.2$\pm$5.0 & \underline{\textbf{36.1$\pm$5.1}} & 47.5$\pm$5.8 & \underline{\textbf{31.8$\pm$7.3}} & \underline{\textbf{25.1$\pm$4.3}} & \underline{\textbf{33.6$\pm$5.7}} \\
 & PGD & 35.0$\pm$8.2 & 23.4$\pm$2.3 & 27.6$\pm$4.5 & 23.6$\pm$5.9 & \textbf{35.5$\pm$6.3} & 35.2$\pm$4.5 & 45.6$\pm$3.0 & 29.7$\pm$8.9 & \underline{\textbf{24.8$\pm$5.1}} & \underline{\textbf{33.3$\pm$7.0}} \\
\midrule \multirow[c]{5}{*}{InputXGrad} & B$_{\text{S}}$ & 31.5$\pm$2.9 & 25.7$\pm$2.8 & 28.2$\pm$1.9 & 21.4$\pm$2.0 & 15.8$\pm$3.4 & 37.6$\pm$3.2 & 52.9$\pm$3.4 & 27.7$\pm$1.7 & 23.1$\pm$1.5 & 31.1$\pm$2.1 \\
 & B$_{\text{U}}$ & 31.5$\pm$2.6 & 24.1$\pm$2.3 & 27.2$\pm$1.2 & 21.0$\pm$1.8 & 20.0$\pm$4.2 & 34.4$\pm$2.6 & 50.9$\pm$3.9 & 25.1$\pm$2.1 & 22.4$\pm$1.1 & 30.0$\pm$1.6 \\
 & IGR & \underline{\textbf{33.7$\pm$2.1}} & \underline{\textbf{24.5$\pm$2.0}} & \underline{\textbf{28.3$\pm$1.1}} & \underline{\textbf{22.5$\pm$1.2}} & 20.6$\pm$3.6 & \underline{\textbf{35.0$\pm$2.8}} & \underline{\textbf{51.7$\pm$3.1}} & \underline{\textbf{26.2$\pm$1.0}} & \underline{\textbf{23.6$\pm$1.2}} & \underline{\textbf{31.5$\pm$1.3}} \\
 & TM & \underline{\textbf{32.2$\pm$4.2}} & 24.1$\pm$1.6 & \underline{\textbf{27.4$\pm$2.0}} & \underline{\textbf{21.2$\pm$3.0}} & \textbf{19.2$\pm$3.3} & \underline{\textbf{34.5$\pm$2.2}} & 50.9$\pm$3.1 & \underline{\textbf{26.3$\pm$2.1}} & \underline{\textbf{22.9$\pm$1.6}} & \underline{\textbf{30.7$\pm$2.3}} \\
 & PGD & 31.2$\pm$2.0 & 24.0$\pm$1.7 & 27.0$\pm$1.2 & 20.3$\pm$1.2 & \textbf{19.7$\pm$4.2} & 34.1$\pm$2.0 & 50.6$\pm$2.4 & \underline{\textbf{25.4$\pm$1.4}} & 22.0$\pm$0.7 & 29.5$\pm$0.9 \\
\midrule \multirow[c]{5}{*}{IG} & B$_{\text{S}}$ & 29.0$\pm$3.0 & 26.9$\pm$1.4 & 27.8$\pm$0.9 & 23.1$\pm$1.6 & 10.7$\pm$4.0 & 41.1$\pm$2.5 & 56.0$\pm$3.1 & 28.7$\pm$2.4 & 23.8$\pm$1.4 & 32.0$\pm$1.9 \\
 & B$_{\text{U}}$ & 31.0$\pm$4.4 & 25.3$\pm$1.9 & 27.8$\pm$2.6 & 22.6$\pm$3.6 & 13.6$\pm$3.2 & 37.7$\pm$2.7 & 53.0$\pm$2.9 & 27.4$\pm$3.2 & 22.8$\pm$2.4 & 30.5$\pm$3.3 \\
 & IGR & \underline{\textbf{32.7$\pm$3.0}} & \underline{\textbf{26.2$\pm$2.5}} & \underline{\textbf{28.9$\pm$1.7}} & \underline{\textbf{24.4$\pm$1.7}} & 14.3$\pm$4.3 & \underline{\textbf{37.9$\pm$2.7}} & \underline{\textbf{53.9$\pm$3.5}} & \underline{\textbf{29.1$\pm$2.7}} & \underline{\textbf{24.2$\pm$1.4}} & \underline{\textbf{32.3$\pm$2.0}} \\
 & TM & \underline{\textbf{31.7$\pm$4.3}} & 24.9$\pm$2.8 & 27.7$\pm$2.4 & \underline{\textbf{22.8$\pm$2.7}} & 14.2$\pm$4.1 & \underline{\textbf{38.1$\pm$2.6}} & 52.7$\pm$3.9 & \underline{\textbf{28.0$\pm$2.9}} & \underline{\textbf{23.3$\pm$1.8}} & \underline{\textbf{31.1$\pm$2.6}} \\
 & PGD & 29.0$\pm$5.4 & \underline{\textbf{25.8$\pm$2.5}} & 27.0$\pm$3.3 & 21.9$\pm$3.9 & \textbf{12.3$\pm$3.9} & 37.5$\pm$2.6 & 52.8$\pm$2.6 & 27.0$\pm$3.4 & 22.5$\pm$2.3 & 30.1$\pm$3.1 \\
\midrule \multirow[c]{5}{*}{Deeplift} & B$_{\text{S}}$ & 29.4$\pm$2.9 & 25.7$\pm$3.1 & 27.2$\pm$1.7 & 20.4$\pm$1.9 & 13.3$\pm$4.0 & 37.6$\pm$3.3 & 54.0$\pm$3.9 & 26.3$\pm$1.8 & 22.9$\pm$1.6 & 30.8$\pm$2.1 \\
 & B$_{\text{U}}$ & 31.1$\pm$2.1 & 22.7$\pm$2.0 & 26.2$\pm$1.8 & 20.0$\pm$2.0 & 20.0$\pm$2.4 & 33.0$\pm$2.1 & 49.7$\pm$2.9 & 24.4$\pm$1.7 & 21.8$\pm$1.2 & 29.2$\pm$1.8 \\
 & IGR & \underline{\textbf{32.2$\pm$1.9}} & \underline{\textbf{24.3$\pm$1.8}} & \underline{\textbf{27.7$\pm$1.4}} & \underline{\textbf{21.6$\pm$1.5}} & \textbf{19.4$\pm$1.7} & \underline{\textbf{34.8$\pm$2.6}} & \underline{\textbf{52.0$\pm$3.2}} & \underline{\textbf{25.6$\pm$1.4}} & \underline{\textbf{22.8$\pm$1.3}} & \underline{\textbf{30.4$\pm$1.5}} \\
 & TM & \underline{29.9$\pm$3.5} & \underline{\textbf{24.5$\pm$1.5}} & \underline{\textbf{26.8$\pm$1.9}} & \underline{\textbf{20.1$\pm$2.9}} & \textbf{16.3$\pm$4.2} & \underline{\textbf{34.7$\pm$2.5}} & \underline{\textbf{51.2$\pm$2.5}} & \underline{\textbf{25.0$\pm$2.0}} & \underline{\textbf{22.3$\pm$1.5}} & \underline{\textbf{29.9$\pm$2.1}} \\
 & PGD & \underline{29.7$\pm$2.6} & \underline{\textbf{23.6$\pm$1.2}} & 26.2$\pm$0.9 & 19.4$\pm$1.0 & \textbf{17.8$\pm$3.7} & \underline{\textbf{33.7$\pm$1.1}} & \underline{\textbf{50.7$\pm$2.1}} & \underline{\textbf{24.5$\pm$1.5}} & 21.4$\pm$0.6 & 28.6$\pm$0.9 \\
\midrule \multirow[c]{5}{*}{AttInGrad} & B$_{\text{S}}$ & 37.7$\pm$2.1 & 37.2$\pm$3.5 & 37.3$\pm$1.9 & 32.4$\pm$2.1 & 3.9$\pm$1.6 & 53.6$\pm$3.5 & 69.7$\pm$2.9 & 40.7$\pm$3.2 & 29.6$\pm$1.4 & 39.8$\pm$2.3 \\
 & B$_{\text{U}}$ & 39.1$\pm$2.7 & 33.1$\pm$3.0 & 35.7$\pm$1.9 & 30.2$\pm$2.1 & 8.9$\pm$3.5 & 46.3$\pm$3.7 & 62.8$\pm$4.5 & 37.2$\pm$2.8 & 28.3$\pm$1.3 & 37.9$\pm$1.5 \\
 & IGR & 36.0$\pm$3.6 & \underline{\textbf{33.3$\pm$2.5}} & 34.6$\pm$2.7 & 28.6$\pm$2.6 & \textbf{7.2$\pm$2.0} & 45.4$\pm$3.8 & 61.3$\pm$4.4 & 36.0$\pm$3.9 & 27.6$\pm$1.8 & 36.9$\pm$2.6 \\
 & TM & \underline{38.7$\pm$3.3} & \underline{\textbf{34.7$\pm$2.9}} & \underline{\textbf{36.4$\pm$1.8}} & \underline{\textbf{30.7$\pm$2.6}} & \textbf{8.4$\pm$4.1} & \underline{\textbf{48.1$\pm$3.3}} & \underline{\textbf{64.3$\pm$3.8}} & 37.2$\pm$1.9 & \underline{\textbf{28.9$\pm$1.5}} & \underline{\textbf{38.6$\pm$1.8}} \\
 & PGD & 34.8$\pm$5.8 & 33.0$\pm$2.3 & 33.5$\pm$3.1 & 26.6$\pm$4.6 & \textbf{7.2$\pm$2.6} & 45.5$\pm$3.1 & 61.3$\pm$2.9 & 32.4$\pm$6.7 & 26.9$\pm$2.9 & 36.1$\pm$4.0 \\
\bottomrule
\end{tabular}
}

\end{table*}

\begin{table*}
\centering
\caption{Plausibility comparison on unsupervised models on the bigger test set. Each experiment was run with ten different seeds. We show the mean of the seeds $\pm$ as the standard deviation. All the scores are presented as percentages. Bold numbers outperform the unsupervised baseline model (B$_{\text{U}}$)}
\label{tab:plausibility_both_uns}
\resizebox{\textwidth}{!}{%
\begin{tabular}{llcccccccccccc}
\toprule
 &  & \multicolumn{7}{c}{Prediction} & \multicolumn{3}{c}{Ranking}  \\
 \cmidrule(lr){3-9} \cmidrule(lr){10-12}
 Explainer& Model & P $\uparrow$ & R $\uparrow$ & F1 $\uparrow$ & AUPRC $\uparrow$ & Empty $\downarrow$ & SpanR $\uparrow$ & Cover $\uparrow$  & IOU $\uparrow$ & P@5 $\uparrow$ & R@5 $\uparrow$  \\ 
 \midrule
\multirow[c]{4}{*}{Rand} & B$_{\text{U}}$ & 0.4$\pm$0.2 & 1.0$\pm$0.8 & 0.4$\pm$0.1 & 0.2$\pm$0.3 & 66.7$\pm$32.6 & 1.2$\pm$0.8 & 2.4$\pm$1.9 & 0.1$\pm$0.0 & 0.1$\pm$0.0 & 0.1$\pm$0.0 \\
 & PGD & 0.4$\pm$0.2 & 1.0$\pm$0.8 & 0.4$\pm$0.1 & 0.5$\pm$0.8 & 66.9$\pm$32.3 & 1.2$\pm$0.8 & 2.5$\pm$2.0 & 0.1$\pm$0.0 & 0.1$\pm$0.0 & 0.1$\pm$0.0 \\
 & TM & 0.4$\pm$0.1 & 1.3$\pm$0.9 & 0.4$\pm$0.0 & 0.6$\pm$0.9 & 60.2$\pm$32.9 & 1.5$\pm$1.0 & 3.0$\pm$2.3 & 0.1$\pm$0.0 & 0.1$\pm$0.0 & 0.1$\pm$0.0 \\
 & IGR & 0.4$\pm$0.2 & 0.8$\pm$0.7 & 0.4$\pm$0.1 & 0.4$\pm$0.6 & 74.0$\pm$30.6 & 0.9$\pm$0.8 & 1.9$\pm$1.9 & 0.1$\pm$0.0 & 0.1$\pm$0.0 & 0.1$\pm$0.0 \\
  \midrule
\multirow[c]{4}{*}{Attention} & B$_{\text{U}}$ & 36.2$\pm$2.5 & 36.8$\pm$4.1 & 36.4$\pm$2.7 & 31.4$\pm$2.9 & 22.2$\pm$3.1 & 52.4$\pm$4.2 & 61.1$\pm$3.4 & 38.1$\pm$2.3 & 29.0$\pm$2.1 & 39.0$\pm$2.8 \\
 & PGD & \textbf{38.5$\pm$2.8} & \textbf{37.3$\pm$2.0} & \textbf{37.7$\pm$1.1} & \textbf{33.1$\pm$1.4} & 22.6$\pm$3.7 & \textbf{52.8$\pm$1.8} & 61.0$\pm$3.4 & \textbf{39.1$\pm$1.4} & \textbf{30.0$\pm$0.8} & \textbf{40.4$\pm$1.1} \\
 & TM & \textbf{37.3$\pm$2.9} & 36.6$\pm$4.2 & \textbf{36.8$\pm$3.0} & \textbf{31.9$\pm$3.2} & 23.5$\pm$3.3 & 52.1$\pm$4.4 & 60.6$\pm$3.7 & \textbf{38.9$\pm$2.7} & \textbf{29.5$\pm$2.2} & \textbf{39.7$\pm$3.0} \\
 & IGR & \textbf{38.0$\pm$1.6} & \textbf{39.1$\pm$1.8} & \textbf{38.5$\pm$1.1} & \textbf{33.5$\pm$1.4} & \textbf{21.2$\pm$1.6} & \textbf{54.5$\pm$1.8} & \textbf{62.1$\pm$1.7} & \textbf{39.6$\pm$1.4} & \textbf{30.4$\pm$0.7} & \textbf{41.0$\pm$0.9} \\
  \midrule
\multirow[c]{4}{*}{Rollout} & B$_{\text{U}}$ & 1.7$\pm$0.0 & 23.8$\pm$1.1 & 3.2$\pm$0.1 & 0.3$\pm$0.0 & 0.0$\pm$0.0 & 24.5$\pm$1.3 & 31.0$\pm$1.5 & 0.4$\pm$0.0 & 0.5$\pm$0.0 & 0.7$\pm$0.0 \\
 & PGD & 1.7$\pm$0.1 & 24.8$\pm$5.9 & 3.1$\pm$0.2 & 0.4$\pm$0.0 & 0.0$\pm$0.0 & 25.4$\pm$6.4 & 31.3$\pm$6.1 & 0.4$\pm$0.0 & 0.5$\pm$0.0 & 0.7$\pm$0.0 \\
 & TM & 1.8$\pm$0.0 & 25.0$\pm$1.1 & 3.3$\pm$0.1 & 0.4$\pm$0.0 & 0.0$\pm$0.0 & 25.8$\pm$1.3 & 32.0$\pm$1.4 & 0.4$\pm$0.0 & 0.5$\pm$0.0 & 0.7$\pm$0.0 \\
 & IGR & 1.7$\pm$0.1 & 25.2$\pm$4.4 & 3.2$\pm$0.2 & 0.4$\pm$0.0 & 0.0$\pm$0.0 & 25.5$\pm$4.7 & 31.7$\pm$4.5 & 0.4$\pm$0.0 & 0.6$\pm$0.0 & 0.8$\pm$0.0 \\
  \midrule
\multirow[c]{4}{*}{$a \nabla a$} & B$_{\text{U}}$ & 35.7$\pm$4.9 & 34.4$\pm$3.7 & 34.9$\pm$3.6 & 29.5$\pm$3.7 & 18.0$\pm$2.2 & 50.2$\pm$4.1 & 61.6$\pm$3.3 & 37.0$\pm$4.4 & 28.5$\pm$3.1 & 38.3$\pm$4.2 \\
 & PGD & \textbf{38.9$\pm$3.3} & \textbf{36.9$\pm$2.4} & \textbf{37.7$\pm$1.4} & \textbf{32.7$\pm$1.9} & \textbf{16.9$\pm$1.9} & \textbf{52.9$\pm$2.5} & \textbf{63.8$\pm$1.4} & \textbf{40.3$\pm$1.8} & \textbf{30.9$\pm$1.1} & \textbf{41.6$\pm$1.5} \\
 & TM & 34.6$\pm$5.3 & \textbf{36.3$\pm$3.5} & \textbf{35.3$\pm$3.9} & \textbf{30.0$\pm$3.9} & \textbf{16.5$\pm$2.5} & \textbf{52.2$\pm$3.8} & \textbf{63.5$\pm$3.1} & \textbf{37.4$\pm$4.5} & \textbf{28.9$\pm$3.2} & \textbf{38.9$\pm$4.3} \\
 & IGR & \textbf{38.1$\pm$3.0} & \textbf{39.1$\pm$2.6} & \textbf{38.4$\pm$1.1} & \textbf{33.2$\pm$1.1} & \textbf{15.7$\pm$1.9} & \textbf{55.5$\pm$2.5} & \textbf{65.5$\pm$2.2} & \textbf{41.2$\pm$1.5} & \textbf{31.3$\pm$0.8} & \textbf{42.2$\pm$1.1} \\
  \midrule
\multirow[c]{4}{*}{InputXGrad} & B$_{\text{U}}$ & 32.4$\pm$2.6 & 31.9$\pm$1.8 & 32.0$\pm$0.8 & 25.7$\pm$1.3 & 10.1$\pm$2.1 & 45.6$\pm$1.9 & 60.8$\pm$1.8 & 31.6$\pm$1.0 & 26.8$\pm$0.6 & 36.0$\pm$0.8 \\
 & PGD & 31.1$\pm$1.7 & 31.7$\pm$2.2 & 31.3$\pm$1.0 & 24.5$\pm$1.6 & 9.7$\pm$1.7 & 45.2$\pm$2.5 & 60.6$\pm$2.8 & 30.7$\pm$1.0 & 26.2$\pm$0.7 & 35.3$\pm$1.0 \\
 & TM & 32.3$\pm$1.9 & \textbf{34.3$\pm$2.1} & \textbf{33.1$\pm$0.9} & \textbf{26.9$\pm$1.1} & 8.8$\pm$1.4 & \textbf{48.3$\pm$2.2} & \textbf{63.5$\pm$2.1} & \textbf{32.7$\pm$0.9} & \textbf{27.5$\pm$0.6} & \textbf{36.9$\pm$0.8} \\
 & IGR & \textbf{33.9$\pm$2.9} & \textbf{33.0$\pm$2.8} & \textbf{33.2$\pm$0.7} & \textbf{27.6$\pm$0.9} & 10.5$\pm$2.7 & \textbf{47.1$\pm$2.9} & \textbf{62.4$\pm$2.9} & \textbf{33.2$\pm$0.9} & \textbf{27.6$\pm$0.5} & \textbf{37.1$\pm$0.7} \\
  \midrule
\multirow[c]{4}{*}{IG} & B$_{\text{U}}$ & 33.0$\pm$2.4 & 32.9$\pm$2.2 & 32.9$\pm$1.7 & 26.6$\pm$3.5 & 5.9$\pm$1.4 & 49.3$\pm$2.6 & 64.3$\pm$2.6 & 34.5$\pm$3.3 & 27.9$\pm$1.1 & 37.5$\pm$1.5 \\
 & PGD & 32.2$\pm$4.5 & 31.8$\pm$1.9 & 31.8$\pm$2.0 & 25.7$\pm$2.6 & \textbf{5.7$\pm$2.1} & 48.0$\pm$2.3 & 63.1$\pm$2.4 & 34.1$\pm$1.8 & 26.9$\pm$1.3 & 36.2$\pm$1.7 \\
 & TM & \textbf{33.8$\pm$2.7} & \textbf{33.6$\pm$2.6} & \textbf{33.6$\pm$1.7} & \textbf{27.7$\pm$3.0} & 5.9$\pm$1.3 & \textbf{50.6$\pm$3.0} & \textbf{65.2$\pm$2.9} & \textbf{35.9$\pm$3.1} & \textbf{28.4$\pm$1.0} & \textbf{38.2$\pm$1.3} \\
 & IGR & \textbf{34.3$\pm$2.5} & \textbf{33.5$\pm$2.2} & \textbf{33.8$\pm$1.5} & \textbf{27.9$\pm$2.4} & 5.9$\pm$1.2 & \textbf{49.6$\pm$2.5} & 64.3$\pm$2.5 & \textbf{35.6$\pm$1.7} & \textbf{28.4$\pm$0.7} & \textbf{38.2$\pm$0.9} \\
  \midrule
\multirow[c]{4}{*}{Deeplift} & B$_{\text{U}}$ & 30.6$\pm$1.5 & 32.5$\pm$1.6 & 31.5$\pm$0.8 & 24.9$\pm$1.4 & 8.3$\pm$1.2 & 46.4$\pm$1.7 & 62.1$\pm$1.4 & 30.7$\pm$0.9 & 26.2$\pm$0.6 & 35.3$\pm$0.8 \\
 & PGD & 30.2$\pm$2.2 & 31.7$\pm$2.8 & 30.8$\pm$0.9 & 24.0$\pm$1.3 & 8.9$\pm$2.1 & 45.3$\pm$3.3 & 61.1$\pm$3.0 & 30.0$\pm$1.0 & 25.8$\pm$0.7 & 34.7$\pm$0.9 \\
 & TM & \textbf{33.2$\pm$1.2} & 32.0$\pm$2.4 & \textbf{32.5$\pm$0.9} & \textbf{26.3$\pm$1.0} & 9.4$\pm$1.5 & 46.1$\pm$2.7 & 62.0$\pm$2.7 & \textbf{31.9$\pm$0.7} & \textbf{27.1$\pm$0.5} & \textbf{36.5$\pm$0.6} \\
 & IGR & \textbf{33.7$\pm$2.3} & 32.0$\pm$2.2 & \textbf{32.7$\pm$0.7} & \textbf{26.8$\pm$1.1} & \textbf{10.3$\pm$2.0} & 46.0$\pm$2.2 & 61.8$\pm$2.6 & \textbf{32.2$\pm$1.1} & \textbf{27.2$\pm$0.6} & \textbf{36.5$\pm$0.8} \\
  \midrule
\multirow[c]{4}{*}{AttInGrad} & B$_{\text{U}}$ & 41.2$\pm$1.1 & 42.1$\pm$3.2 & 41.5$\pm$1.6 & 37.0$\pm$1.6 & 2.8$\pm$0.6 & 59.1$\pm$3.0 & 73.5$\pm$2.0 & 43.2$\pm$1.3 & 32.7$\pm$1.1 & 44.0$\pm$1.5 \\
 & PGD & 39.7$\pm$1.8 & \textbf{44.0$\pm$2.5} & \textbf{41.7$\pm$0.8} & \textbf{37.2$\pm$0.9} & \textbf{2.3$\pm$0.5} & \textbf{60.8$\pm$2.3} & \textbf{74.4$\pm$1.8} & \textbf{43.3$\pm$0.7} & \textbf{32.9$\pm$0.4} & \textbf{44.2$\pm$0.6} \\
 & TM & \textbf{41.4$\pm$1.7} & \textbf{43.2$\pm$3.1} & \textbf{42.2$\pm$1.7} & \textbf{37.7$\pm$1.8} & \textbf{2.7$\pm$0.6} & \textbf{60.3$\pm$2.8} & \textbf{74.8$\pm$2.1} & \textbf{43.7$\pm$1.4} & \textbf{33.0$\pm$1.1} & \textbf{44.4$\pm$1.5} \\
 & IGR & 40.9$\pm$2.3 & \textbf{43.9$\pm$2.5} & \textbf{42.2$\pm$0.5} & \textbf{38.1$\pm$0.6} & \textbf{2.5$\pm$0.8} & \textbf{60.8$\pm$2.2} & \textbf{74.2$\pm$2.1} & \textbf{44.1$\pm$0.5} & \textbf{33.2$\pm$0.4} & \textbf{44.7$\pm$0.5} \\
\bottomrule
\end{tabular}}
\end{table*}

\end{document}